\documentclass{article}

     \usepackage[preprint]{neurips_2025}

\usepackage[utf8]{inputenc} 
\usepackage[T1]{fontenc}    
\usepackage{hyperref}       
\usepackage{url}            
\usepackage{booktabs}       
\usepackage{amsfonts}       
\usepackage{nicefrac}       
\usepackage{microtype}      
\usepackage{xcolor}         
\usepackage{wrapfig}   
\usepackage{amssymb}

\usepackage{multirow}
\usepackage{caption}
\usepackage{amsmath}
\usepackage{amsthm}
\usepackage{dsfont}
\usepackage{algorithm}
\usepackage{algorithmic}
\usepackage[pdftex]{graphicx}
\usepackage{threeparttable}
\usepackage{subcaption}
\usepackage{mathtools}
\usepackage{placeins}
\usepackage{adjustbox}

\theoremstyle{plain}
\newtheorem{theorem}{Theorem}[section]
\newtheorem{proposition}[theorem]{Proposition}
\newtheorem{lemma}[theorem]{Lemma}

\theoremstyle{definition}

\newtheorem{assumption}[theorem]{Assumption}
\newtheorem{example}[theorem]{Example}
\theoremstyle{remark}

\newcommand\cM{\mathcal M}
\newcommand\cL{\mathcal L}

\newcommand\cD{\mathcal D}

\newcommand\bbR{\mathbb R}
\newcommand\bbE{\mathbb E}
\newcommand\bbN{\mathbb N}
\newcommand\sY{\mathsf Y}
\newcommand\sTheta{\mathsf \Theta}

\title{Comparison of parallel SMC and MCMC for Bayesian deep learning}

\author{%
  Xinzhu Liang\\
  Mathematics Department, University of Manchester\\
  Manchester, M13 9PL, UK \\
  \texttt{xinzhu.liang@postgrad.manchester.ac.uk} \\
  \And
  Joseph M. Lukens\thanks{Quantum Information Science Section, Oak Ridge National Laboratory, Oak Ridge, Tennessee 37831, USA} \\
  School of Electrical and Computer Engineering\\
  Purdue University, West Lafayette, Indiana 47907, USA\\
  \AND
  Sanjaya Lohani \\
  Department of Electrical and Computer Engineering, Southern Methodist University \\
  Dallas, Texas 75205, USA\\
  \And
  Brian T. Kirby\thanks{Tulane University, New Orleans, Louisiana 70118, USA} \\
  DEVCOM US Army Research Laboratory\\ Adelphi, Maryland 20783, USA\\
  \And
  Thomas~A. Searles \\
  Department of Electrical and Computer Engineering, University of Illinois Chicago\\
  Chicago, Illinois 60607, USA\\
  \And
  Xin Qiu \\
  Cognizant AI Labs\\
  San Francisco, California 94105, USA
  \And
  Kody J.~H. Law \\
  Mathematics Department, University of Manchester\\
  Manchester, M13 9PL, United Kingdom\\
}

\begin{document}

\maketitle

\begin{abstract}
This work systematically compares 
parallel implementations of 
{\em consistent} (asymptotically unbiased) Bayesian deep learning algorithms:
sequential Monte Carlo sampler (SMC$_\parallel$) or Markov chain Monte Carlo (MCMC$_\parallel$).
We provide a proof of convergence for SMC$_\parallel$
showing that it theoretically achieves the same level of convergence as a single monolithic SMC sampler,
while the reduced communication lowers wall-clock time.
It is well-known that the first samples from MCMC need to be discarded to eliminate initialization bias,
and that the number of discarded samples must grow like the logarithm of the number of parallel chains
to control that bias for MCMC$_\parallel$.
A systematic empirical numerical study on MNIST, CIFAR, and IMDb, reveals that 
parallel implementations of both methods perform comparably to
non-parallel implementations in terms of performance and total cost, 
and also comparably to each other.
However, both methods still require a large wall-clock time, 
and suffer from catastrophic non-convergence if they aren't run for long enough.
\end{abstract}

\section{Introduction}
\label{sec:intro}

\setlength{\tabcolsep}{2pt}


Quantification of uncertainty (UQ) in deep learning is critical for 
safe and reliable deployment, yet remains a core challenge.
The Bayesian formulation provides UQ in addition to Bayes optimal accuracy, by averaging 
realizations from the posterior distribution, rather than relying on a single point estimator.
%
Fully Bayesian approaches like consistent 
Markov chain Monte Carlo (MCMC) and sequential Monte Carlo (SMC)
offer asymptotically unbiased posterior estimates, 
but at the cost of prohibitive compute time compared to simple point estimators like the maximum a posteriori (MAP).
Bayesian deep learning (BDL) often rely on scalable approximations
such as Monte Carlo Dropout \citep{gal2016dropout}, 
stochastic variational inference \citep{hoffman2013stochastic}, 
and deep ensembles (DE) \citep{lakshminarayanan2017simple},
which are fast and provide strong empirical performance,
but lack formal consistency guarantees.


Given data \(\cD\), the Bayesian posterior distribution over \(\theta \in \sTheta \in \bbR^{d}\) is given by 
  $  \pi(\theta) \propto \cL(\theta) \pi_0(\theta)$,
where \(\cL(\theta) := \cL(\theta; \cD)\) is the likelihood of
the data \(\cD\) and \(\pi_0(\theta)\) is the prior. 
The Bayes estimator of a quantity of interest \(\varphi:\sTheta \rightarrow \bbR\) is \(\bbE[\varphi|\cD]=\int_\sTheta \varphi(\theta) \pi(\theta) d\theta\). It minimizes the appropriate 
Bayes risk at the population level and as such is
Bayes optimal \citep{mackay1992practical, 
neal2012bayesian, 
andrieu2003introduction, bishop2006pattern}.
 
\begin{wrapfigure}[36]{r}{0.5\textwidth}
        \vspace{-10pt}
        \centering
        \includegraphics[width=.49\columnwidth]{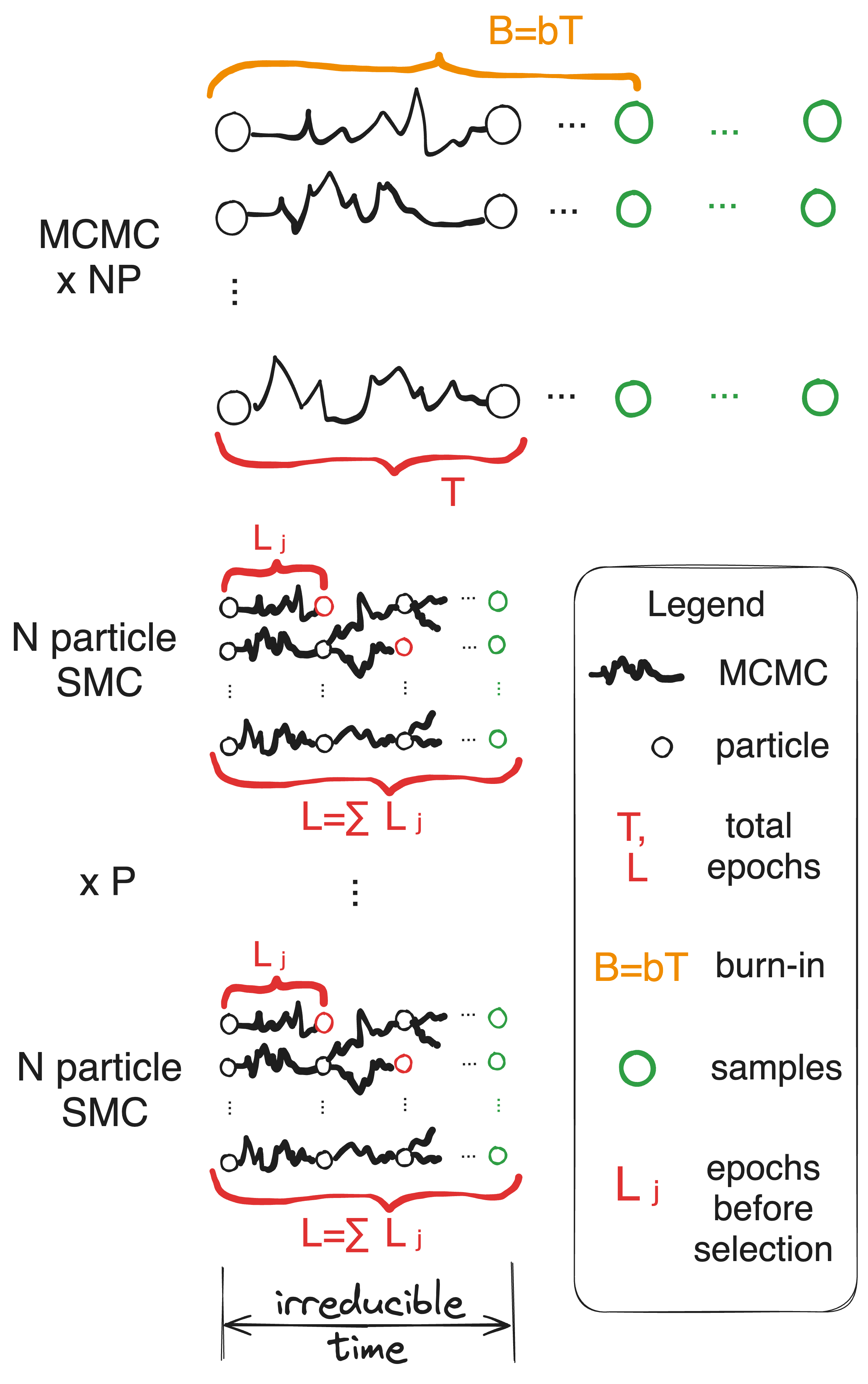}
        \caption{Cartoon diagram of ensembles generated by parallel SMC and MCMC methods.
        Both methods need to be run for sufficiently many total epochs
        over the training data (full likelihood calculations), 
        and are otherwise able to scale in parallel similarly to other ensemble methods.}
        \label{fig:sbmc_cartoon}
\end{wrapfigure}
In general the posterior (target) distribution can only be evaluated up-to a constant of proportionality, and the available consistent methods for inference (learning) are of Monte Carlo type: 
notably Markov chain Monte Carlo (MCMC) 
\citep{metropolis1953equation, hastings1970monte,duane1987hybrid,gelfand1990sampling,geyer1992practical,robert1999monte,roberts1996exponential}
and sequential Monte Carlo (SMC) samplers \citep{del2006sequential,dai2022invitation,chopin2020introduction}.
The past several decades have seen enormous progress in methodology as well as practical applications
\citep{galison2022first,mohan2024evaluating}, 
however standard implementations of these algorithms are still too expensive for practical 
BDL, and 
 consistent Monte Carlo (MC) methods are typically 
 used only as a benchmark for cheaper approximations \citep{izmailov2021bayesian}.
 See e.g. \citep{angelino2016patterns,papamarkou2024position} 
 for recent reviews and further references.
The present work aims to address the computational intractability head-on by exploring consistent MC
methods which can distribute the workload across arbitrarily many workers in parallel,
hence delivering practicality by virtue of scalability.
The focus of the present work is on studying parallelism, and 
we will not consider stochastic gradient MCMC methods 
\cite{welling2011bayesian,chen2014stochastic} nor the virtues of mini-batching
or other data-parallel techniques 
in terms of scalability and convergence.
However, we note that these and other more sophisticated adaptive 
methods can be swapped in later for additional gains.

MCMC methods originated with the famous Metropolis-Hastings (MH) methods
\citep{metropolis1953equation, hastings1970monte}, and are the favoured approach to consistently approximate this kind of target distribution in general. MCMC has seen widespread use and rigorous development in statistics from the turn of the millennium
\citep{duane1987hybrid,gelfand1990sampling,geyer1992practical,robert1999monte,roberts1996exponential}.
The sequential Monte Carlo (SMC) sampler 
\citep{del2006sequential,dai2022invitation,chopin2020introduction} 
is an alternative population-based MC method, which was developed at the turn of the millennium \citep{jarzynski1997equilibrium,berzuini2001resample,gilks2001following,neal2001annealed,chopin2002sequential}. 
The SMC sampler approximates the target distribution with a
population of sample ``particles'' 
which evolve through importance re-sampling (selection)
and MCMC moves (mutation) between 
intermediate
distributions.  

The majority of parallel MCMC methods 
leverage communicating parallel chains primarily for the purpose of improved mixing and convergence 
 \citep{martino2016orthogonal, craiu2009learn, earl2005parallel,brockwell2006parallel,chen2016accelerated, calderhead2014general, schwedes2021rao,syed2022non}.
The idea of simply combining multiple {\em independent} chains 
(or batches of communicating parallel chains) 
is appealing, but due to the serial nature of the method it
is only recently beginning to gain widespread attention 
\citep{wilkinson2006parallel,chen2016accelerated,jacob2020unbiased,margossian2022modernizing,margossian2023many,margossian2024nested,de2022parallel,sountsov2024running,hoffman2019langevin,hoffman2021adaptive,hoffman2020black,hoffman2022tuning,nguyen2025unorthodox}, 
and the approach has seen limited application to BDL.
The upshot is that this approach 
typically suffers from a bias like \(\exp(-b)/N^2\), 
if each chain runs for $NT$ steps after 
$bT$ warm-up samples are discarded,  
with mixing time $T$
\citep{robert1999monte,roberts1996exponential,betancourt2021short,margossian2024nested}.
We will denote this method as MCMC$_\parallel$.

SMC is inherently parallel, as the 
MCMC moves for each particle can be executed concurrently 
\citep{lee2010utility,paige2014asynchronous,syed2024optimised}.
This is where the likelihood computations happen, which are the most expensive and dominant contributors
to the computational complexity.
One must take care that sufficient memory per core is available
or else this approach will hit the ``memory wall'' \citep{ivanov2021data}.
The resampling stage requires communication between all particles, which could also be a bottleneck 
depending on the communication bandwidth, 
but this may not be an issue for {Synchronous Single Instruction, Multiple Data} (SIMD)
architectures as found 
on a single multi-core CPU or GPU.
This is the most common approach for leveraging parallelism in SMC, 
as it is very simple to implement and can sometimes deliver strong parallel scaling \citep{lee2010utility}.

Beyond this intra-SMC parallelism, the combination of parallel SMCs
has recently 
been considered in \cite{verge2015parallel, whiteley2016role}. 
The recommended approach 
typically involves communication between {\em all samples}, 
which improves stability but hinders scalability for large models 
which needs to be distributed 
across 
{many SIMD nodes},
which may either be disconnected or have slower between-node inter-connect. 
The island particle model \citep{verge2015parallel} separates the total number of samples \(N_{\sf total}\) 
into \(P \leq P_{\rm max}\) 
SMC islands with \(N\) samples each. 
Without between-SMC interaction, naive (equal-weight) averaging results in 
an asymptotic bias penalty \((1/N)^2\) \citep{crisan2018performance,verge2015parallel}.
However, {\em by assigning appropriate weights to each SMC 
we can eliminate this penalty} \citep{whiteley2016role,dai2022invitation}.
The resulting scalable parallel SMC sampler is denoted by SMC$_\parallel$.

The contributions of the present work are concisely summarized as follows:
\begin{itemize}
\item[(a)] A theoretical 
complexity result for 
SMC$_\parallel$ is given in Theorem \ref{thm:psmc_converge};
\item[(b)] Systematic numerical experiments 
show:
\begin{itemize}
\item 
(i) comparable performance between MCMC$_\parallel$ and SMC$_\parallel$
{\em provided both methods are run for long enough}.
\item 
(ii) catastrophic breakdown of MCMC$_\parallel$ and SMC$_\parallel$ 
on first order metrics 
if the chains are not run for long enough.
\end{itemize}
\end{itemize}

\section{Setup and Algorithms}
\label{sec:setup}

\textbf{The MCMC algorithm} is given in Algorithm \ref{alg:mcmc}. The serial implementation is the basic version of the MCMC algorithm, and the parallel implementation features the naive $N$ parallel short chains free from any communication. Let the MCMC transition kernel be \(\cM\), 
such that \((\pi \cM)(d\theta) = \pi(d\theta)\).
We will employ two standard MCMC kernels: preconditioned Crank-Nicolson (pCN) 
\citep{bernardo1998regression,cotter2013mcmc}
and Hamiltonian Monte Carlo (HMC) \citep{duane1987hybrid, neal2011mcmc}.
Details 
are given in Appendix \ref{app:mcmc}. 

\begin{wrapfigure}[15]{r}{0.45\textwidth}
  \vspace{-10pt}
  \begin{minipage}{\linewidth}
  \begin{algorithm}[H]
   \caption{MCMC}
   \label{alg:mcmc}
\begin{algorithmic}
    \STATE \textbf{Inputs:} $\cL$, $\pi_0$, $N$.
    \STATE Initialise: If serial, \(\theta^1_0 \sim \pi_0\); if parallel, \(\theta^i_0 \sim \pi_0\) for $i=1,...,N$. $J=B$.
    \FOR{\(i=1\) {\bfseries to} \(N\) (serial or parallel)} 
    \STATE If serial, $\theta^{i+1}_0\leftarrow \theta^i_{J}$; $J=T$.
    \FOR{\(j=1\) {\bfseries to} \(J\) (in serial)}
    \STATE Draw \(\theta_{j}^i \sim \cM(\theta_{j-1}^i, \cdot )\).
    \ENDFOR
    \ENDFOR
    \STATE \textbf{Outputs:} $\{\theta^i=\theta^i_{J}\}_{i=1}^{N}$ (and $Z^N\equiv 1$\footnote{Defined for consistency of notation 
    in \eqref{eq:pmc}.}).
\end{algorithmic}
\end{algorithm}
\end{minipage}
\end{wrapfigure}
\textbf{The SMC sampler algorithm} \citep{del2004feynman} 
alternates between {\em selection} by importance re-sampling,
and {\em mutation} according to an appropriate intermediate MCMC transition kernel.  
Define a sequence of intermediate targets $\pi_j(\theta) \propto \cL(\theta)^{\lambda_j}\pi_0(\theta)\, ,$ according to a tempering schedule $0=\lambda_0,\dots,\lambda_J=1$, which will be chosen adaptively according to the effective sample as described in \ref{ex:adaptive_temp} in the Appendix.
The intermediate MCMC transition kernel, \(\cM_{j}\), is defined 
such that \((\pi_{j} \cM_{j})(d\theta) = \pi_{j}(d\theta)\) and 
$\cM_{J}(d\theta)=\cM(d\theta)$ \citep{geyer1992practical}.
This operation must sufficiently decorrelate the samples, 
and as such we 
define the MCMC kernels \({\cM_{j}}\) by several 
steps of the basic pCN or HMC kernel,
leading to 
$L_j$
likelihood/gradient evaluations, 
or {\em epochs}. 
In the case of HMC kernel, there are also 
several leapfrog steps for each HMC step contributing to $L_j$. See Algorithm \ref{alg:smc}.

For a quantity of interest \(\varphi: \sTheta \rightarrow \bbR\), the MCMC or SMC estimator from 
Algorithm \ref{alg:mcmc} 
or \ref{alg:smc} is 
\begin{equation}
\pi^N(\varphi)  :=  \frac{1}{N} \sum_{i=1}^{N} \varphi(\theta^{i}) \, .
\end{equation}
\textbf{The MCMC$_\parallel$ algorithm} refers to $P$ parallel executions of Algorithm \ref{alg:mcmc} with $N_{\sf total}=NP$ and parallel $N-$loop as default. 

\begin{wrapfigure}[16]{l}{0.5\textwidth}
  \vspace{-25pt}
  \begin{minipage}{\linewidth}
  \begin{algorithm}[H]
    \caption{SMC}
    \label{alg:smc}
    \begin{algorithmic}
       \STATE \textbf{Inputs:} $\cL$, $\pi_0$, $N$.
       \STATE Init. \(\theta^i_0 \sim \pi_0\) for \(i=1,\dots,N\). \(Z^N=1\). 
       \FOR{\(j=1\) {\bfseries to} \(J\) (in serial)}
         \STATE (Optional) Select 
         $\lambda_j$ s.t. ESS$=\alpha N$.
         \STATE Store \(Z^N *= 
         \frac{1}{N} \sum_{k=1}^N \cL(\theta_{j-1}^k)^{\lambda_j-\lambda_{j-1}}\).
         \vspace{3pt}
         \FOR{\(i=1\) {\bfseries to} \(N\) (in parallel)}
             \vspace{2pt}
            \STATE Define \(w^i_j \propto \cL(\theta_{j-1}^i)^{\lambda_j-\lambda_{j-1}}\).
           \STATE {\bf Selection}: 
           \(I_j^i \sim \{w_j^1,\dots,w_j^N\}\).
           \STATE {\bf Mutation}: 
           \(\theta_j^i \sim \cM_j(\theta_{j-1}^{I_j^i},
           \cdot ~)\).
         \ENDFOR
       \ENDFOR
       \STATE \textbf{Outputs:} $\{\theta^{i}=\theta_{J}^{i}\}_{i=1}^{N}$ and $Z^{N}$.
    \end{algorithmic}
  \end{algorithm}
\end{minipage}
\end{wrapfigure}
\textbf{The SMC$_\parallel$ algorithm} 
refers to $P$ parallel executions of Algorithm \ref{alg:smc} with \(N\) particles each,
for \(N_{\sf total}=NP\) samples in total.
It has a \(P\) times lower 
communication and memory overhead than a single monolithic SMC sampler with \(N_{\sf total}\) particles. 
This simplification 
is crucial for massive problems such as BDL, which require distributed architectures. SIMD resources can be used for the $N$ communicating particles
(and model- and data-parallel likelihood calculations within individual epochs), 
while minimizing and optimally utilizing scarce interconnected resources.

The parallel estimator resulting from Algorithm \ref{alg:psmc} is given by
     \begin{equation}\label{eq:pmc}
     \hat{\varphi}= \sum_{p=1}^{P} \omega_p \pi^{N,p}(\varphi) \, , \quad \omega_p =\frac{Z^{N,p}}{\sum_{p=1}^P Z^{N,p}} \, .
     \end{equation}

\begin{wrapfigure}[10]{r}{0.4\textwidth}
 \vspace{-20pt}
\begin{minipage}{0.35\textwidth}
\begin{algorithm}[H]
    \caption{
    SMC$_\parallel$ and MCMC$_\parallel$}
    \label{alg:psmc}
\begin{algorithmic}
    \STATE \textbf{Inputs:} $\cL$, $\pi_0$, $N$.
    \FOR{\(p=1\) {\bfseries to} \(P\) (in parallel)}
    \STATE Run Algorithm~\ref{alg:smc} (SMC) or \ref{alg:mcmc} (MCMC) on $\pi$.
    \STATE Output \(\{\theta^{i,p}\}_{i=1}^N\) and \(Z^{N,p}\)\,.
    \ENDFOR
       \STATE \textbf{Outputs:} $\{\{\theta^{i,p}\}_{i=1}^{N},Z^{N,p}\}_{p=1}^P$.
\end{algorithmic}
\end{algorithm}
\end{minipage}
\end{wrapfigure}

\begin{figure}
    \hfill
    \includegraphics[width=0.45\columnwidth]{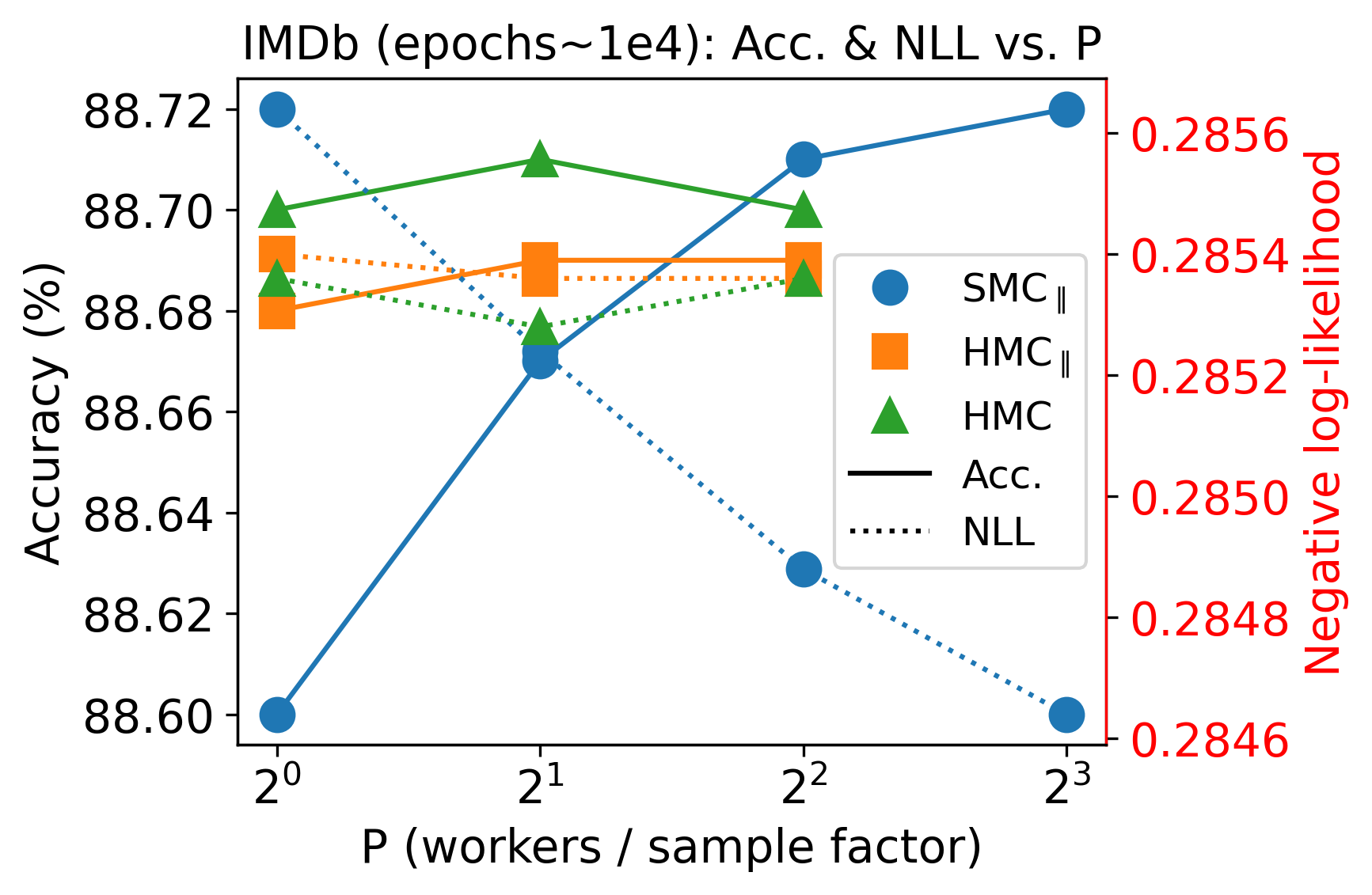} 
    \hfill
    \includegraphics[width=0.45\columnwidth]{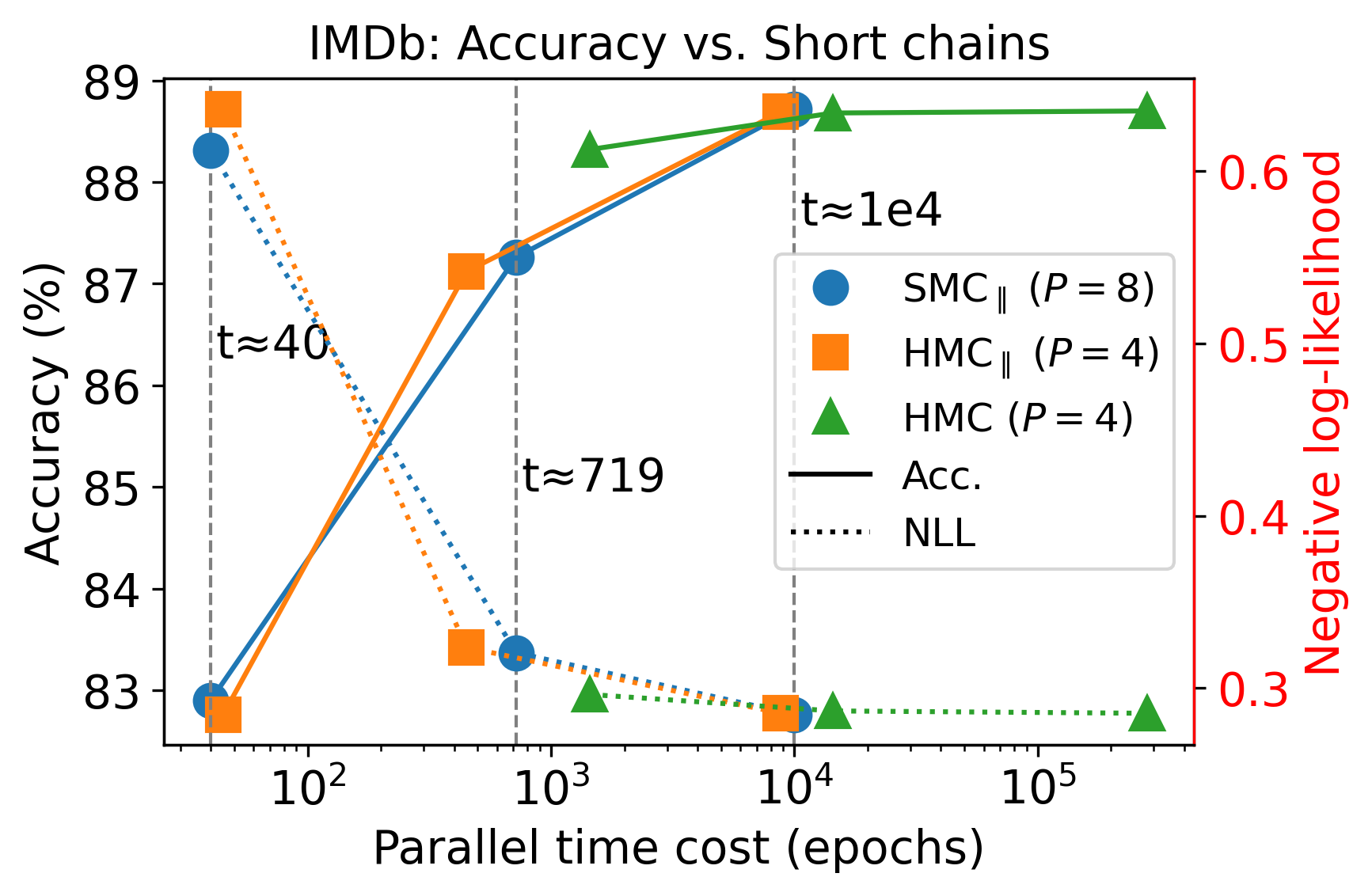}   
    \caption{Left: test accuracy and NLL over $P$ for serial HMC 
    (single chain run for $TNP$ steps, with the first $T=1e4$ steps discarded as burn-in), 
    HMC$_\parallel$ (a single sample from each of $NP$ chains run for $T$ burn-in steps),
    and SMC$_\parallel$ ($P$ SMC with $N$ interacting particles each), 
    with $N=32$.
    Right: converged (in $P$) values showing catastrophic failure for short chains.}
\label{fig:punchline-s1}
\vspace{-15pt}
\end{figure}

A concise synopsis of the method is presented in Figure \ref{fig:punchline-s1}
for the IMDb sentiment classification dataset\footnote{https://huggingface.co/datasets/stanfordnlp/imdb} 
    \citep{maas2011}.  
The left panel shows performance with respect to $P$ for sufficiently long wall-clock time for individual processes, 
while the right panel shows catastrophic non-convergence when the individual processes are not run for long enough
($P=8$). Note that we use epochs, i.e. likelihood plus gradient evaluation as a hardware-agnostic proxy to measure wall-clock time per processor. 
These can be further parallelized with model- and data-parallel techniques. 




\section{Comparison of SMC and MCMC methods}
\label{sec:smc-mcmc}

Note, in this section, we consider $P=1$ always and hence $N=N_{\sf total}$.

\textbf{Serial Implementation. }
Let $C$ denote the cost to evaluate the likelihood and its gradient, i.e. one {\em epoch}.
Suppose MCMC delivers correlated samples with an integrated autocorrelation time (IACT\footnote{IACT is defined in Appendix \ref{app:IACT}.}) of 
$T_{A}$
epochs.
Discarding $bT_{A}$
epochs as warm-up, after 
$NT_{A}$ more epochs, 
the total computational complexity is $(b+N-1)T_{A}C$ and the error is expected to be 
MSE$=\mathcal{O}(\exp(-b)/N^2+1/N)$ \citep{robert1999monte,roberts1996exponential,betancourt2021short,margossian2024nested}. 
For SMC, the total computational complexity to evolve $N$ samples through $J$ tempering stages
with $L_j$ epochs
each time is $LNC$ for an error MSE$=\mathcal{O}(1/N)$ \citep{del2004feynman}, 
where $L=\sum_{j=1}^J L_j$. 
This is equivalent to MCMC if $T_{A}=L$ and $b=1$.
Note the SMC constant hidden in $\mathcal{O}$ can become prohibitively large for 
small $L$.
In practice $M$ and $J$ are selected adaptively and we find $L\approx T_{A}$. 
\setlength{\tabcolsep}{5pt}



\textbf{Intra-parallel Implementation. } 
If suitable hardware is available, 
SMC allows parallelization of the mutation steps 
over $N$ cores with
identical results for a {\em time complexity} of $CL+NJ$ 
(ordinarily the second term should be much smaller),
ignoring potential slow-down due to communication, which should be minimal 
{\em provided this is intra-node SIMD-style communication}.
Also, the discussion above implies we can retain a single sample 
from each of $N$ parallel MCMCs 
and achieve MSE$=\mathcal{O}(\exp(-b)+1/N)$ \citep{margossian2024nested}, 
which would be indistinguishable from the single chain result for appropriate 
$b=\mathcal{O}(\log N)$. 
The parallel cost is also equivalent if 
$T_A=L$.
However, 
the MCMC estimator is not consistent for finite $b$, 
and this can potentially spoil convergence.

\begin{figure}[H]
    \centering
    \includegraphics[width=.9\columnwidth]{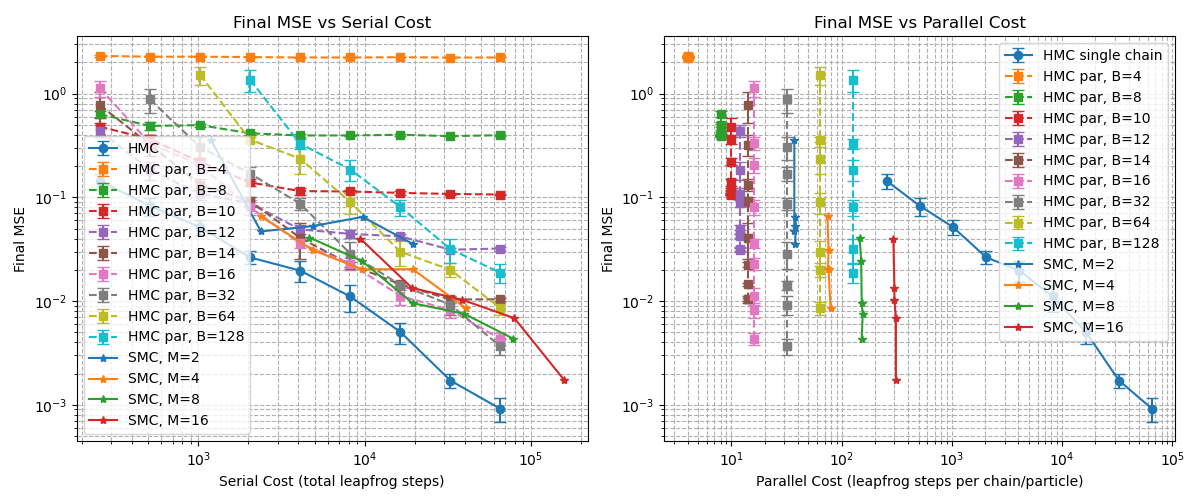} \\
    \caption{MSE vs. serial 
    and parallel cost of HMC and SMC for various 
    parameter settings, for a tractable Gaussian example.
    $M$ is the number of mutation steps for MCMC. For the parallel implementation of 
    HMC (parallelized in $N$, and distinct from what we denote as HMC$_\parallel$),
    a single sample is taken from $N/B$ chains of length $B$,
    for varying $N$.}
\label{fig:singletons}
\end{figure}
\begin{figure}[H]
    \centering
    \includegraphics[width=.9\columnwidth]{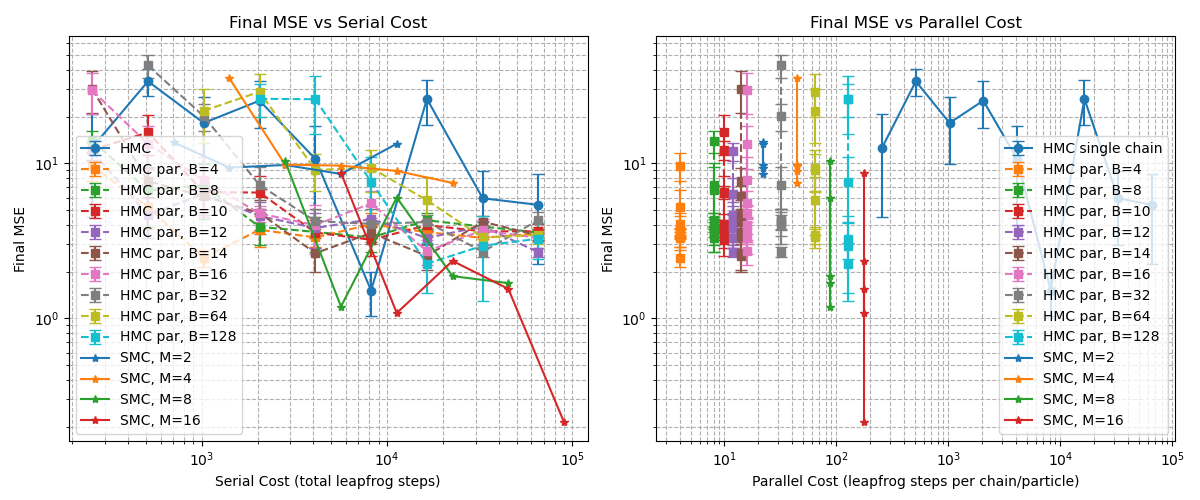}    
    \caption{The same as Figure \ref{fig:singletons} for 
    a tractable GMM: MSE vs. serial 
    and parallel cost of HMC and SMC for various 
    parameter settings, for a tractable Gaussian example.}
\label{fig:singletons2}
\end{figure}
{\bf Examples. } Figures \ref{fig:singletons}-\ref{fig:singletons2} show the MSE vs serial cost 
(left: $N-$loop not executed in parallel in Algorithms \ref{alg:smc} and \ref{alg:mcmc}) 
and parallel wall-clock cost (right) 
of HMC and SMC (with HMC mutations) 
for various parameter settings and two problems.
Figures \ref{fig:singletons} corresponds to a tractable Gaussian posterior
over $\theta \sim \mathcal{N}(0,{\sf Id})$
with $d=16$ and 
$m=32$ observations of $y\sim \mathcal{N}(X {\bf 1}_d,{\sf Id})$, with $X_{ij}\sim \mathcal{N}(0,1)$.
Figures \ref{fig:singletons2} corresponds to a tractable Gaussian mixture model (GMM) 
$0.2\mathcal{N}({\bf 1}_d, {\sf Id}) + 0.8\mathcal{N}(-{\bf 1}_d, {\sf Id})$.
A standard normal initialization is used for all.
    There is a small overhead in total serial (computational) cost for the parallel methods,
    but this pays dividends in total parallel (wall-clock time) cost.
    $M$ is the number of HMC mutation steps for SMC, with $10$ leapfrog steps per HMC step,
    so $L_j=10M$ for all $j$,
    and the step size is adapted between tempering steps to target an optimal acceptance rate of $0.65$ \citep{beskos2013optimal},
    for number of particles $N \in \{32, 64, \dots
    512\}$.
    For HMC the step size is also adapted to target $0.65$ acceptance rate,
    and total number of leapfrog steps $L_{\sf total} \in \{256, 512, \dots,
    65536\}$. For $B \in \{4,8,\dots,128\}$,
    a single sample is taken from $N/B$ chains of length $B$,
    where $N=\lceil L_{\sf total}/10 \rceil$ is the total number of HMC steps.
    For the Gaussian example in \ref{fig:singletons}, 
    single HMC is the most efficient in total computation,
    while parallel implementations show significant improvement 
    in parallel cost for large enough $B$.
    For $B<16$, the HMC bias is apparent.
    SMC with $M=2$ also 
    stagnates.
    For the GMM example in \ref{fig:singletons2}, 
    only SMC converges, for $M=16$.
    SMC is 
    more robust to multi-modality and initialization, 
    but can also struggle in high-dimension
    \citep{buchholz2021adaptive}.



\textbf{Free parameters } for MCMC can be selected with standard techniques 
\citep{beskos2013optimal,cotter2013mcmc,carpenter2017stan,buchholz2021adaptive}.
SMC additionally requires tuning of $M$ and tempering schedule,
which can both also be done adaptively \citep{dai2022invitation}.
We found $N\geq N_{\sf min}$ can be quite small in practice, 
e.g. 16 or 32. 
In principle, any MCMC adaptation scheme can be used within SMC
although 
approaches which leverage the particle population,
such as \cite{gilks1994adaptive,vrugt2009accelerating,
hoffman2022tuning},
make more sense than 
serial approaches like \cite{haario2001adaptive,hoffman2014no}.
See also Appendix \ref{app:free}. 

\section{Parallel SMC (SMC$_\parallel$) and MCMC (MCMC$_\parallel$) methods}
\label{sec:psmc}

\begin{figure}
    \centering
    \includegraphics[width=1\columnwidth]{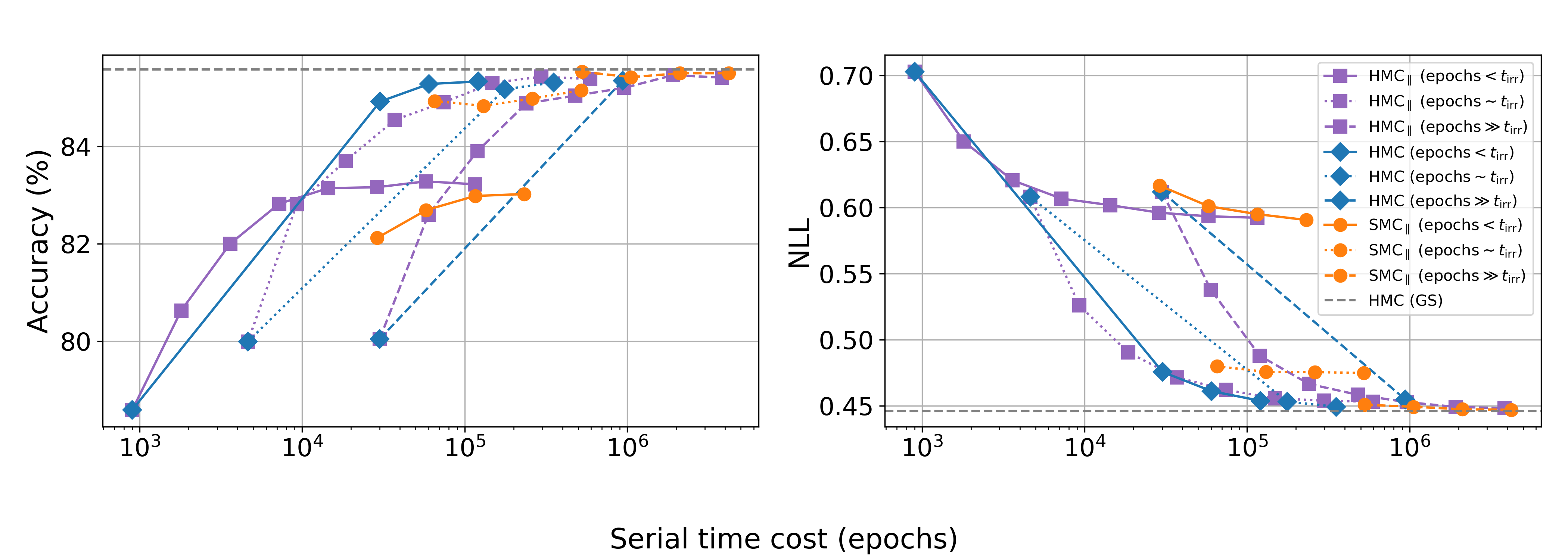}
    \caption{MNIST with
    SMC\(_\parallel\) (\(N=32\) each), HMC\(_\parallel\) 
    ($B=900$ steps each),
    and single serial HMC,  
    for different fixed trajectory lengths.} 
  \label{fig:mnist_metric_fixtraj-main}
\end{figure}
We now look more closely at SMC$_\parallel$ and MCMC$_\parallel$.
Define $t_{\sf irr}=1e4$, which is the number of epochs that 
need to be run in serial for the methods to converge, 
i.e. this limits how small the wall-clock time can be.
Figure \ref{fig:mnist_metric_fixtraj-main}
shows SMC\(_\parallel\) (\(N=32\) each, $20$ steps per mutation), 
HMC\(_\parallel\) ($B=900$ steps each), and single serial HMC,  
    for different per-step fixed trajectory lengths 
    $\tau=0.005$ (epochs $<t_{\sf irr}$),
    $\tau=0.02$ (epochs $\lesssim t_{\sf irr}$), and $\tau=0.1$ (epochs $\gg t_{\sf irr}$),
    over $P$ (MNIST).
    The plot clearly shows 
    (i) substantial improvements in HMC$_\parallel$ from $1$ up to $32\times P$ parallel chains,
    (ii) comparable performance of SMC$_\parallel$ and HMC$_\parallel$ at $32$ chains,
    and (iii) convergence to sub-optimal plateau for epochs $<t_{\sf irr}$
    and GS
    for epochs $>t_{\sf irr}$.



\textbf{A theoretical complexity result for SMC$_\parallel$} is now presented.
It requires only standard assumptions,
which essentially state that 
the likelihood is bounded above and below and 
the MCMC kernel is strongly mixing. 
The precise assumptions and proof 
are given in the Appendix \ref{app:theorySec}.

\begin{theorem} 
\label{thm:psmc_converge}
    Given Assumptions \ref{ass:like_bound} and \ref{ass:mutation}, 
    for suitable 
    \(\varphi, M, N, J\), 
    there exists a \(C
    >0\), which depends on \(\varphi,M,J\), such that for any \(P \in \bbN\),
    $\bbE[(\hat{\varphi}_{\text{SMC$_\parallel$}} - \pi(\varphi))^2] \leq \frac{C}
    {NP}.$
\end{theorem}
    
    




\begin{wrapfigure}[24]{l}{0.5\textwidth}
\vspace{-20pt}
    \centering
    \footnotesize
    \includegraphics[width=0.5\columnwidth]{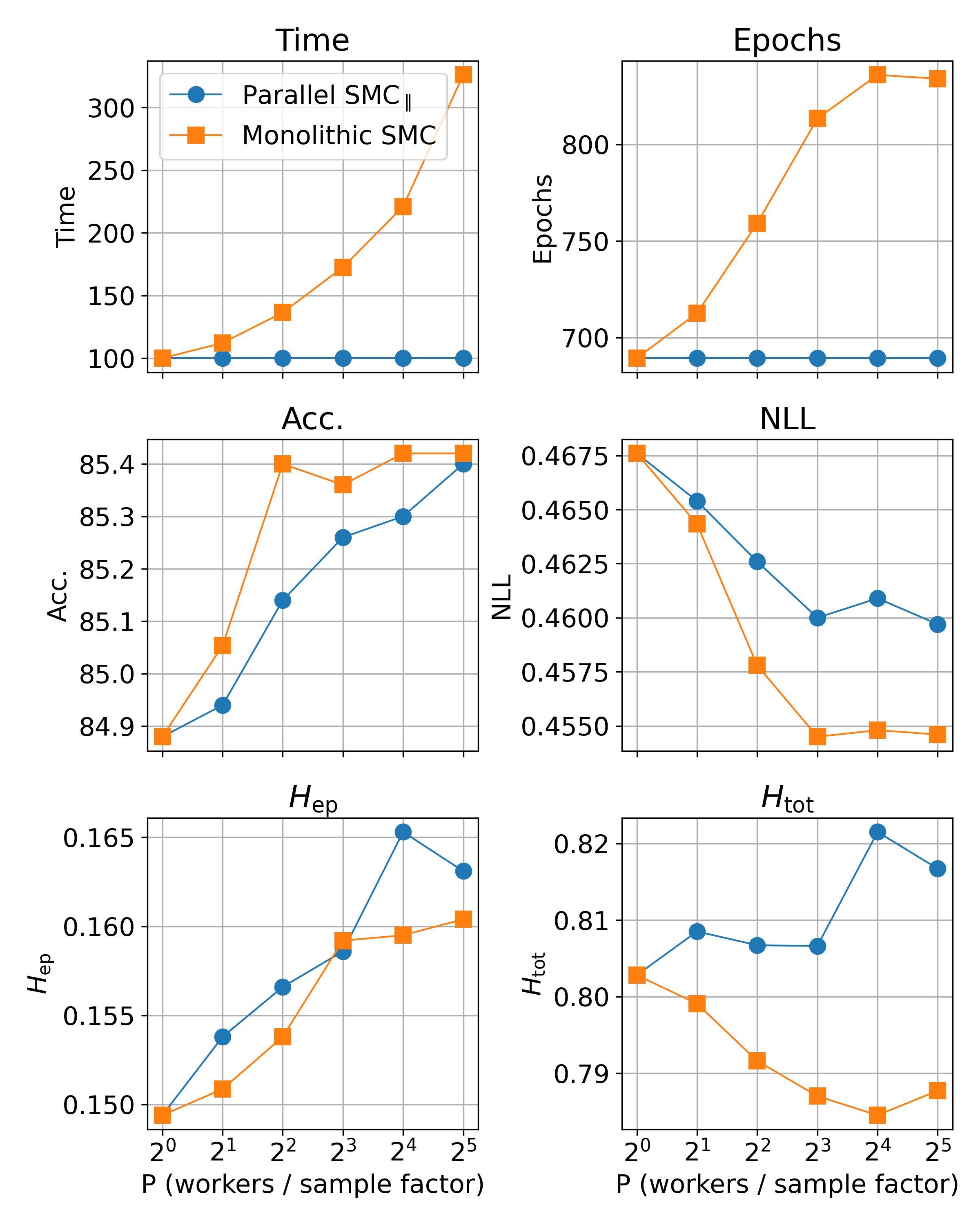}
    \caption{SMC$_\parallel$ 
    \& SMC.
    $P$ nodes, $N-$cores each.}
\label{fig:nvp}
\end{wrapfigure}
\textbf{Communication overhead of interconnect and $N \leftrightarrow P$ exchangeability. }
We compare a single monolithic SMC with SMC$_\parallel$ on 
MNIST, whose architecture is
described in Appendix \ref{app:mnist}. 
Recall that 
SMC is already \(N\)-parallel {\em with communication}, 
whereas the $P$-parallelism 
is {\em free from any communication}.
SMC$_\parallel$ with $P$ SMC of $N=32$ particles each
is compared to a single monolithic 
SMC with $PN$ particles 
on our special inter-connected {\em HPC Pool} of $32$ nodes with $32$ cores each.
From the upper left panel, we can observe a factor $P$ slowdown due to communication.
From the upper right panel, the single SMC requires more epochs as $P$ increases,
up to $20\%$ for $P=16$, but most slowdown is communication-related.
This is consistent with the observation that intra-node communication 
is usually very efficient,
as opposed to between-node inter-connect \cite{ivanov2021data}.
Note that in a separate experiment  
on a single node,
for $32$ vs. $1$ core we observe
a speed-up factor of $\approx 8$ in a single chain/particle 
(Appendix {\ref{app:interconnect}} Table \ref{tab:mnist_hpc_differcore}), in comparison to
a factor $\approx 32$ speed-up for $32$ chains/particles. 
We do however see a small slowdown
in epochs/s from $\approx 10$ to $\approx 8.5$ for SMC, 
due to {\em intra-node} communication overhead.
See also Appendix \ref{app:interconnect} for expanded data,
\ref{app:par_gauss} for large $P$ on a Gaussian model,
and \ref{app:exchange} for 
more examples of $N \leftrightarrow P$ exchangeability for various 
wall-clock times.
\textbf{SMC$_{\parallel}$ and MCMC$_{\parallel}$ vs SOTA on Australian Credit data \citep{statlog_(australian_credit_approval)_143} comparison. }
\begin{figure}
\centering
\includegraphics[width=0.7\textwidth]{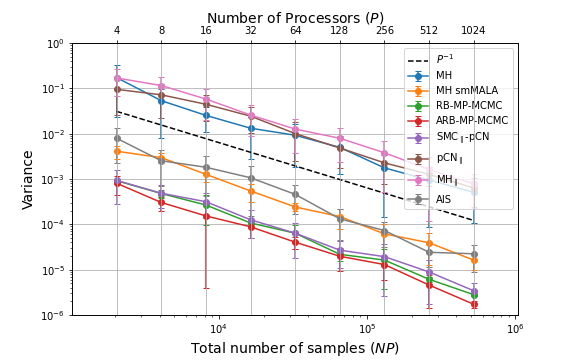}
 \caption{Variance of SMC$_\parallel$-pCN, various MCMC$_\parallel$, 
 and AIS, over total samples ($NP$) on the top axis and number of 
 parallel processes ($P$) on the bottom.}
 \label{fig:parallelMCMC_AIS}
\end{figure}
The dataset has $d=15$ covariates and $m=690$ data and
Bayesian logistic regression is used.
We compare SMC$_\parallel$-PCN with annealed importance sampling 
(AIS--equivalent to $N=1$ particle per process) \citep{neal2001annealed}, 
various MCMC$_\parallel$ methods, and  
recent synchronous-parallel MCMC \citep{schwedes2021rao}.
Further details are provided in the Appendix \ref{app:other}.
Figure~\ref{fig:parallelMCMC_AIS} 
shows the empirical variance for posterior mean estimates of the proposed methods with respect to the number of samples
(bottom axis) and number of parallel processes (top axis),
verifying the theoretical $1/P$ convergence result of Theorem \ref{thm:psmc_converge}.
Data from \cite{schwedes2021rao} is copied directly.
SMC$_\parallel$ 
performs comparably to the others.
Among methods which are free from communication along $P$, our method performs the best, 
outperforming the MCMC$_\parallel$ methods and AIS.
{SMC and AIS incur a per-sample overhead in cost ($L$), 
but this is offset by intra-parallelism ($L/N \ll 1$).}
\textbf{SMC$_\parallel$ and MCMC$_\parallel$ in a nutshell (theory and practice). }
{As shown in Theorem \ref{thm:psmc_converge}, 
the SMC$_\parallel$ estimator with $P$ samplers
of fixed size $N\geq N_{\rm min}$ 
converges 
as \(P\rightarrow \infty\), and without any loss of efficiency: 
with \(P\) {\em non-interacting} processors, the method converges 
at the rate MSE\(=\mathcal{O}(1/P)\)
with \(\mathcal{O}(1)\) time complexity. 
However, the constant becomes prohibitively large when chains are too short or
$N<N_{\rm min}$,
leading to a large shelf in practice (see Appendix \ref{app:free}).
In contrast, MCMC$_\parallel$ chains of length $(b+N-1)T$ will have the rate MSE$=\mathcal{O}(\exp(-b)/N^2 + 1/NP)$,
which means that $b$ should be chosen proportionally to $\log(P)$. 
Hence MCMC$_\parallel$ has a $\mathcal{O}(\log(P))$ complexity, and it is 
the same asymptotic bias which can 
leads to non-convergence in practice for small $b$.
See 
Appendix \ref{app:const_trick} for further implementation details.
}

It is important to note that the error discussed above is
 $\varepsilon(\hat{y},\hat{y}_{\sf BAYES})$, where $\hat{y}_{\sf BAYES}$ is the Bayes
optimal posterior predictive estimator and $\hat{y}$ is prediction from the model. 
A practical consideration is that one is often primarily concerned with the 
generalisation errors 
$\varepsilon(\hat{y},y) = \varepsilon(\hat{y},\hat{y}_{\sf BAYES}) + 
{\varepsilon(\hat{y}_{\sf BAYES},y)},$
so asymptotic results are most 
valuable in the infinite data limit when 
$\varepsilon(\hat{y}_{\sf BAYES},y)=\mathcal{O}(1/\sqrt{m})\rightarrow 0$, at which point UQ is of limited value 
due to Bernstein von-Mises Theorem \citep{van2000asymptotic,sullivan2015introduction}.

\section{Further Directions}
\label{sec:further}


Further directions include
\begin{itemize}
\item $P-$parallelizing $N-$ensemble MCMC methods such as \cite{gilks1994adaptive,goodman2010ensemble,vrugt2009accelerating,hoffman2022tuning}.
\item Leveraging $N-$ensemble MCMC methods within SMC for better mutations (with the cost of more communication).
\item Parallel stochastic-gradient-MCMC methods like SGLD \citep{welling2011bayesian} and SG-HMC \citep{chen2014stochastic}, 
and ensemblized versions thereof.
\item Related to above, mini-batch gradients can be used in lieu of full 
gradients, which may have some advantages in terms of scalability and convergence.
For SMC samplers, we have unbiased estimators $\widehat{\ell w}$ of log weights using mini-batches, and could use $\exp(\widehat{\ell w})$ for a non-negative and biased estimator or
Bernoulli/Poisson augmentation to achieve (non-negative) unbiased weights \citep{gunawan2021subsamplingSMC,vihola2020poissonSMC}.
\end{itemize}

\subsection{Discussion of Limitations}
\label{sec:limits}


We have shown that SMC$_\parallel$ and MCMC$_\parallel$ both work in practice,
i.e. both methods are parallel. However, the individual chains need to be run for 
a long time, which is the most significant limitation to scalability.
The largest problems 
require distributing individual particles across multiple cores
and leveraging 
data parallel likelihood computation
\citep{izmailov2021bayesian,grattafiori2024llama}, 
as well as 
model-parallel approaches 
 \citep{huggingface_parallelism}.
The best conceivable complexity is $\mathcal{O}(d)$ for memory (per particle), 
$\mathcal{O}(dL)/\mathcal{O}(dbT)$ for time, and $\mathcal{O}(dLN)/\mathcal{O}(d(b+N)T)$ for compute (SMC/MCMC). 
Ideally we would have guarantees for 
$N,L_j,J,T$ all 
constant in $d$, which seems conceivable from practical experiments. 

Our sampler relies only on forward/back‑prop evaluations, so every mainstream hardware scheme can be stacked on top of it: data‑parallel all‑reduce for moderate models\citep{goyal2017imagenet1h}; 
optimizer‑state sharding (ZeRO/FSDP) when 
parameters no longer fit\citep{rajbhandari2020zero}; 
tensor model‑parallelism for in‑layer splits \citep{shoeybi2019megatron} and pipeline model‑parallelism for depthwise splits \citep{huang2019gpipe}; 
and, finally, the full hybrid of DP/sharding/tensor/pipeline that is now routine in trillion‑parameter language models \citep{chowdhery2022palm,black2022gptneox}.

\section{Conclusion}
\label{sec:conclusion}

Both MCMC$_\parallel$ and SMC$_\parallel$ are attractive algorithm options,
which are consistent for the posterior provided they are run for long enough.
Therefore, they both provide a ``gold-standard'' baseline for BDL
against which to evaluate other methods, given a suitable budget.
However, {\em we would not recommend running them far-from convergence},
as performance degrades catastrophically.
SMC has more attractive theoretical properties, 
such as $\mathcal{O}(1)$ vs. $\mathcal{O}(\log P)$ scaling for $P$ processes, 
while MCMC has benefits in practical application, 
such as communication-free parallelism.

\begin{ack}

KJHL and XL gratefully acknowledge the support of IBM and EPSRC in the form of an Industrial Case Doctoral Studentship Award. 
JML acknowledges funding from the U.S. Department of Energy, Office of Science, Advanced Scientific Computing Research (Early Career Research Program, ReACT-QISE).

\end{ack}

\bibliography{refs}




\appendix

\section{MCMC kernels}
\label{app:mcmc}

The specific MCMC kernels used are presented here.

\subsection{Pre-conditioned Crank-Nicolson kernel}

The original pCN kernel was introduced in \cite{bernardo1998regression, cotter2013mcmc}.
The general pCN kernel is given by
\begin{equation*}
    \theta^{\prime} = \Sigma_0^{1/2}({\sf Id} - \beta^2 D)^{1/2} \Sigma_0^{-1/2} (\theta-\mu_0) + \mu_0 + \Sigma_0^{1/2} \beta D^{1/2} \delta \, ,
\end{equation*}
where the proposal in the original/standard pCN is with \(D = {\sf Id}\). The general version presented above was introduced in \cite{law2014proposals, cui2016dimension}, and can provide substantially improved mixing when the likelihood informs certain directions much more than others. The scaling matrix \(D\) should be chosen according to the information present in the likelihood. A simple and computationally convenient choice, which is particularly amenable to use within SMC, is to build it from an approximation of the target covariance \cite{beskos2018multilevel}. In the present work, we adopt the simplest and cheapest choice and
let \(D = \text{diag}(\widehat{\text{var}}[\theta])\). The approximation of the variance, \(\widehat{\text{var}}\), will be built from the current population of samples in the SMC case, 
and adaptively constructed in the MCMC case \cite{haario2001adaptive, chen2016accelerated}. See also \cite{rudolf2018generalization,beskos2014stable,zahm2022certified}.

\begin{algorithm}[h!]
    \caption{pCN kernel}
    \label{alg:pre_cn_kernel}
\begin{algorithmic}
    \STATE \textbf{Inputs:} a current state \(\theta\), the distribution \(\pi_j\), a scaling parameter \(\beta\).
    \FOR{\(m=1\) {\bfseries to} \(M\)}
    \STATE Generate \(\theta^{\prime} = \Sigma_0^{1/2}({\sf Id}-\beta^2 D)^{1/2} \Sigma_0^{-1/2} (\theta-\mu_0) + \mu_0 + \Sigma_0^{1/2} \beta D^{1/2} \delta\) where \(\delta \sim \mathcal{N}(0,{\sf Id})\) and \(u \sim U([0,1])\).
    \IF{\(u \leq \min \Big\{ 1,\frac{\cL(\theta^{\prime})^{\lambda_j}}{\cL(\theta)^{\lambda_j}} \Big\}\)}
    \STATE \(\theta = \theta^{\prime}\).
    \ELSE{}
    \STATE \(\theta = \theta\).
    \ENDIF
    \ENDFOR
    \STATE \textbf{Outputs:} \(\theta\).
\end{algorithmic}
\end{algorithm}

\subsection{Hamiltonian Monte Carlo}

Hamiltonian Monte Carlo (HMC) kernel \cite{duane1987hybrid, neal2011mcmc, houlsby2011bayesian, cobb2021scaling} is 
essentially a gradient-based MCMC kernel on an extended state-space. We first build a Hamiltonian \(H(\theta, q)\) with additional auxiliary ``momentum'' vector \(q\) of the same dimension as the state \(\theta\)
\begin{equation}
    H(\theta,q) = -\log \pi_j(\theta) + \frac{1}{2} q^{\top} M_0^{-1} q,
\end{equation}
where \(M_0\) is a mass matrix, so that \(\frac1Z \exp(-H(\theta,q))\) is a target distribution on the extended space, where the momentum can be simulated exactly \(q\sim \mathcal{N}(0,M_0)\).

From physics, we know that the Hamiltonian dynamics conserve energy, hence 
avoiding local random-walk type behaviour and allowing ballistic moves in position:

\begin{equation}
\begin{aligned}
    & \frac{d \theta}{dt} = \frac{\partial H}{\partial q} = M_0^{-1} q, \\
    & \frac{d q}{d t} = \frac{\partial H}{\partial \theta} = \nabla_{\theta} \log \pi_j(\theta) \, .
\end{aligned}
\end{equation}
A carefully constructed symplectic-integrator is capable of approximately conserving energy as well, for example
the leapfrog integrator \cite{leimkuhler2004simulating}:
\begin{equation}
\begin{aligned}
\label{eqn:leapfrog_step}
    & q_{t+\Delta t/2} = q_{t} + \frac{\Delta t}{2} \frac{dq}{dt}(\theta_t),\\
    & \theta_{t+\Delta t} = \theta_t + \Delta t \frac{d\theta}{dt}(q_{t+\Delta t/2}), \\
    & q_{t+\Delta t} = q_{t+\Delta t/2} + \frac{\Delta t}{2} \frac{dq}{dt}(\theta_{t+\Delta t}),
\end{aligned}
\end{equation}
where \(t\) is the leapfrog step iteration and \(\Delta t\) is the step size. 

Each step of the HMC method is summarized as follows:
\begin{itemize}
\item[(i)] simulate a random momentum \(q\sim \mathcal{N}(0,M_0)\) (hence jumping to a new energy contour);
\item[(ii)] approximate the Hamiltonian dynamics using \(L\) steps from \eqref{eqn:leapfrog_step};
\item[(iii)] correct numerical error from (ii) with an MH accept/reject step for \(\frac1Z \exp(-H(\theta,q))\).
\end{itemize}
See Algorithm \ref{alg:hmc_kernel}.

\begin{algorithm}[h!]
    \caption{Hamiltonian Monte Carlo kernel}
    \label{alg:hmc_kernel}
\begin{algorithmic}
    \STATE \textbf{Inputs:} a current state \(\theta\), the distribution \(\pi_j\) and a mass matrix \(M_0\).
    \STATE Init. Generate a initial momentum \(q \sim \mathcal{N}(0,M_0)\).
    \FOR{\(m=1\) {\bfseries to} \(M\)}
    \FOR{\(l=1\) {\bfseries to} \(L\)}
    \STATE Generate \(\theta_{l\Delta t}\) and \(q_{l\Delta t}\) from \eqref{eqn:leapfrog_step} with \(\theta_{0} = \theta\) and \(q_{0} = q\)
    \ENDFOR{}
    \STATE Let \(\theta^{\prime} = \theta_{L \Delta t}\), \(q^{\prime} = q_{L \Delta t}\) and generate \(u \sim U([0,1])\)
    \IF{\(u \leq \min \Big\{ 1,\exp(H(\theta,q)-H(\theta^{\prime},q^{\prime})) \Big\}\)}
    \STATE \(\theta = \theta^{\prime}\) and \(q=q^{\prime}\).
    \ELSE{}
    \STATE \(\theta = \theta\) and \(q = q\).
    \ENDIF
    \ENDFOR{}
    \STATE \textbf{Outputs:} \(\theta\).
\end{algorithmic}
\end{algorithm}

\section{Techniques for SMC$_\parallel$ in practice}
\label{app:const_trick}

{\bf Adaptive tempering. } As mentioned, adaptive tempering is used to ensure a dense tempering regime and 
provide stability\cite{syed2024optimised}. 

\begin{example}[Adaptive tempering]
\label{ex:adaptive_temp}
    In order to keep the sufficient diversity of sample population, we let the effective sample size to be at least \(\text{ESS}_{\min} = N/2\) at each tempering \(\lambda_{j-1}\) and use it compute the next tempering \(\lambda_j\). For \(j\)th tempering, we have weight samples \(\{ w^{k}_{j-1}, \theta^{k}_{j-1} \}_{k=1}^{N}\), then the ESS is computed by
    \begin{equation*}
        \text{ESS} = \frac{1}{\sum_{k=1}^{N} (w^{k}_{j-1})^2},
    \end{equation*}
    where \(w^{k}_{j-1} = \cL(\theta^{k}_{j-1})^{\lambda_{j} - \lambda_{j-1}}/\sum_{k=1}^N \cL(\theta_{j-1}^k)^{\lambda_j-\lambda_{j-1}}\). Let \(h = \lambda_j - \lambda_{j-1}\), the effective sample size can be presented as a function of \(h\), ESS\((h)\). Using suitable root finding method, one can find \(h^{*}\) such that \(ESS(h^{*})=ESS_{\min}\), then set the next tempering \(\lambda_j = \lambda_{j-1}+h^{*}\).
\end{example}

{Note that the partition function estimator \(Z^N\) is no longer unbiased
once we introduce adaptation, which means that in principle we should do short pilot runs and 
then keep everything fixed to preserve the integrity of the theory, 
but we have found this does not make a difference in practice.

{\bf Numerical stability: nested Log-sum-exp. } When computing likelihoods in Sequential Monte Carlo (SMC) algorithms, numerical underflow frequently arises because likelihood values can become extremely small, often beyond computational precision. To address this, one standard practice is to work with log-likelihoods rather than likelihoods directly. By operating in the log domain, the computer can safely store and manipulate extremely small values without loss of precision.

Specifically, the standard \emph{log-sum-exp} trick can be applied to stabilize computations. For instance, consider a scenario with nested sums and products in parallel SMC. For each processor \( p = 1, \dots, P \), we initially have:
\[
Z^{N,p} = \prod_{j=1}^{J}\sum_{i=1}^{N}\omega_{j}^{i,p}.
\]

To avoid numerical instability, each sum within the product is computed using the log-sum-exp trick:
\[
\sum_{i=1}^{N}\omega_{j}^{i,p} = \exp\left(\max_{i}\log(w_j^{i,p})\right)\sum_{i=1}^{N}\exp\left(\log(w_j^{i,p}) - \max_{i}\log(w_j^{i,p})\right).
\]

This procedure yields the decomposition:
\[
Z^{N,p} = K^p \hat{Z}^p,
\]
where
\[
K^p = \prod_{j=1}^{J}\exp\left(\max_{i}\log(w_j^{i,p})\right), \quad \text{and} \quad \hat{Z}^p = \prod_{j=1}^{J}\sum_{i=1}^{N}\exp\left(\log(w_j^{i,p}) - \max_{i}\log(w_j^{i,p})\right).
\]

In parallel SMC, an additional stabilization step is applied across processors. The global normalization constant across processors can also suffer from numerical instability. To address this, the log-sum-exp trick is applied again at the processor level:
\[
Z^{N,p} = \exp\left(\log(\hat{Z}^p) + \log(K^p) - \log(K)\right)K,
\]
with
\[
\log(K) = \max_{p}\left(\log(\hat{Z}^p) + \log(K^p)\right).
\]

Since the factor \( K \) cancels out when calculating the parallel SMC estimator, it suffices to compute only:
\[
\exp\left(\log(\hat{Z}^p) + \log(K^p) - \log(K)\right),
\]
which ensures numerical stability even when \(K\) itself is computationally very small.

Thus, by recursively applying the log-sum-exp trick at both the particle and processor levels, parallel SMC estimators can robustly handle computations involving extremely small numbers without numerical underflow.

\section{Theoretical results}
\label{app:theorySec}

Below are the theoretical results for SMC$_\parallel$ complexity of $\mathcal{O}(1)$ in $P$.
We defer to \cite{roberts1996exponential,robert1999monte,margossian2024nested}
for MCMC $\mathcal{O}(\log P)$ complexity.
Note that badly designed SMC$_\parallel$ (unweighted average)
or MCMC$_\parallel$ (fixed $b$) would require $N \propto \sqrt{P}$ and 
have a complexity $\mathcal{O}(\sqrt{P})$.

First, we give a more detailed description of the method. Define 
the target distribution as 
\(\pi(\theta) = f(\theta)/Z\), where \(Z = \int_{\sTheta} f(\theta) d\theta\)
and \(f(\theta):= \cL(\theta) \pi_0(\theta)\). 
For \(j=1,...,J\), we let \(f_j\) define an annealing scheme for \(0=\lambda_0<\lambda_1<\cdots<\lambda_J=1\):
\begin{equation*}
    f_j = f_0^{1-\lambda_j}f_J^{\lambda_j}=\cL(\theta)^{\lambda_j}\pi_0(\theta)\, .
\end{equation*}
Define the successive importance weights 
by \(h_j = f_{j+1}/f_j\), where \(f_0 = \pi_0\) and \(f_{J} = f\),
and define \(\pi_j = f_j/Z_j\) where \(Z_j = \int_{\sTheta} f_j \). 
\(\lambda_j\) will be chosen adaptively according to the effective sample size (ESS), 
as described in 
\ref{ex:adaptive_temp} in the Appendix.

Now let \(\cM_{j}\) for \(j=1,\dots,J\) be any suitable MCMC transition kernels such that \((\pi_{j} \cM_{j})(d\theta) = \pi_{j}(d\theta)\) \citep{geyer1992practical,robert1999monte}. 
This operation must sufficiently decorrelate the samples, 
and as such we typically define the MCMC kernels \({\cM_{j}}\) by several 
steps of some basic MCMC kernel, leading to some number $L_j$
of likelihood/gradient evaluations, which we refer to as {\em epochs}.
In the present work, we employ two standard MCMC kernels: preconditioned Crank-Nicolson (pCN) 
\citep{bernardo1998regression,cotter2013mcmc,law2014proposals}
and Hamiltonian Monte Carlo (HMC) \citep{duane1987hybrid, neal2011mcmc}.
In the latter case, there are also 
several leapfrog steps for each HMC step contributing to $L_j$.
Details 
are given in Appendix \ref{app:mcmc}.

Given a quantity of interest \(\varphi: \sTheta \rightarrow \bbR\), 
the SMC
estimator of $\pi(\varphi)$ and unbiased estimator of $f(\varphi)$
from Algorithm \ref{alg:smc}
are given by 
\begin{equation}
\label{eqn:est_smc}
\pi_J^N(\varphi)  =  \frac{1}{N} \sum_{i=1}^{N} \varphi(\theta_{j}^{i}) \, , 
\quad 
f_J^N(\varphi) = Z_J^N \pi_J^N(\varphi)\, , \quad {\rm where} \quad 
Z_J^N = \prod_{j=1}^{J-1} \pi_j^N(h_j) \, . 
\end{equation}

By separating \(NP\) samples into \(P\) processors with \(N\) samples in each, 
SMC$_\parallel$ has a \(P\) times lower 
communication and memory overhead than a single SMC sampler. 
This simplification 
is crucial for massive problems such as BDL, which require distributed architectures.
Synchronous SIMD style resources can be used for the $N$ communicating particles,
and model- and data-parallel 
likelihood calculations within individual epochs beyond that, while 
minimizing and optimally utilizing scarce interconnect resources.
Algorithm \ref{alg:psmc} displays the SMC$_\parallel$ method.

Following from Algorithm \ref{alg:psmc} and \eqref{eqn:est_smc}, 
we can define the consistent (in $P$, for finite $N$ suitably large)
SMC$_\parallel$ ratio estimator of
\(\pi(\varphi)\),
in terms of the unbiased 
{\em $P-$level un-normalized estimator
}, 
as follows
\begin{align}\label{eq:est_psmc}
\setlength\abovedisplayskip{5pt}%
 \setlength\belowdisplayskip{5pt}%
    \hat{\varphi}_{\text{SMC$_\parallel$}} = \frac{ F^{N,P}(\varphi)}{ F^{N,P}(1)} 
    = \sum_{p=1}^{P} \omega_p \pi^{N,p}(\varphi) \, , \quad \omega_p \propto Z^{N,p} \, ,
    \quad 
        F^{N,P}(\varphi) &= \frac{1}{P} \sum_{p=1}^{P} f^{N,p}(\varphi)\, .
\end{align}

MCMC$_\parallel$ here means
$P$ parallel executions of
Algorithm \ref{alg:mcmc}, 
with parallel $N-$loop as default. 

\subsection{Assumptions, Proposition and Lemma}
\label{app:theory}

We first present the assumptions.

\begin{assumption}
\label{ass:like_bound}
    Let \(J \in \bbN\) be given, 
    there exists a \(C >0\) such that for all \(\theta \in \sTheta\) and \(j \in \{ 1,...,J \}\),
    \begin{equation*}
        C^{-1} < \ f_{j}(\theta), \ \cL(\theta) \leq C.
    \end{equation*}
\end{assumption}

\begin{assumption}
\label{ass:mutation}
    Let \(J \in \bbN\) be given, 
    there exists a \(\rho \in (0,1)\) such that for any \((u,v) \in \sTheta^2\), measurable \(A \in \sTheta\), 
    and \(j \in \{ 1,...,J\}\),
    \begin{equation*}
        \int_{A} \cM_{j}(u,du^{\prime}) \geq \rho \int_{A} \cM_{j}(v,dv^{\prime}).
    \end{equation*}
\end{assumption}

In order to make use of \eqref{eq:est_psmc}, we require estimates on \(f^{N,p}(\zeta)\) both for quantity of interest \(\zeta=\varphi\) and \(\zeta=1\). We denote that \(| \zeta |_{\infty} = \max_{\theta \in \sTheta} |\zeta(\theta)|\) in the following equations. \(C_{\zeta}\) denotes the constant depended on the function \(\zeta\). Note that using \(\varphi\) with one-dimensional output is without loss of generality for our convergence results, and the following proof can be directly generalized to the multi-output function by using the inner product. Proof of the following proposition can be found in \cite{del2004feynman}.

\begin{proposition}
\label{prop:single_smc_converge}
    Assume Assumption \ref{ass:like_bound} and \ref{ass:mutation}. Then, for any \(J \in \bbN\), there exists a \(C > 0\) such that for any \(N \in \bbN\), suitable \(\zeta: \sTheta \rightarrow \bbR\),
    \begin{equation}
        \bbE [ (f^{N}(\zeta) - f(\zeta) )^2 ] \leq  \frac{C |\zeta|_{\infty}^2}{N}.
    \end{equation}
    In addition, the estimator is unbiased \(\bbE[f^{N}(\zeta)] = f(\zeta)\).
\end{proposition}

The following supporting Lemma will be proven in the next section, along with the main theorem \ref{thm:psmc_converge}.

\begin{lemma}
\label{lem:psmc_single}
    Assume Assumption \ref{ass:like_bound} and \ref{ass:mutation}. Then, for any \(J \in \bbN\), there is a \(C > 0\) such that for suitable $N$ and \(\zeta:\sTheta \rightarrow \bbR\), and any \(P \in \bbN\),
    \begin{equation*}
        \bbE[(F^{N,P}(\zeta) - f(\zeta))^2] \leq \frac{C |\zeta|_{\infty}^2 }{NP}.
    \end{equation*}
\end{lemma}

\subsection{Proofs}
\label{app:proofs}

The proofs of the various results in the paper are presented here, along with restatements of the results.

\subsection{Proof relating to Lemma \ref{lem:psmc_single}}
\label{app:lemma_psmc_single}

{\em Assume Assumption \ref{ass:like_bound} and \ref{ass:mutation}. Then, for any \(J \in \bbN\), there is a \(C > 0\) such that for any \(N,P \in \bbN\), suitable \(\zeta:\sTheta \rightarrow \bbR\), 
\begin{equation*}
\bbE[(F^{N,P}(\zeta) - f(\zeta))^2] \leq \frac{C |\zeta|_{\infty}^2 }{NP}.
\end{equation*}}
\begin{proof}
\begin{align}
    &\bbE [(F^{N,P}(\zeta) - f(\zeta))^2] \nonumber\\ 
    &= \bbE \bigg[ \bigg(\frac{1}{P} \sum_{p=1}^{P} (f^{N,r}(\zeta) - f(\zeta)) \bigg)^2 \bigg] \nonumber \\
    &= \frac{1}{P^2} \bbE \bigg[ \sum_{p=1}^{P}(f^{N,p}(\zeta)-f(\zeta))^2 + \sum_{p=1}^{P} \sum_{p^{\prime}=1}^{P} (f^{N,p}(\zeta)-f(\zeta)) (f^{N,p^{\prime}}(\zeta)-f(\zeta)) \bigg] \nonumber\\
    &= \frac{1}{P^2} \sum_{p=1}^{P} \bbE [ (f^{N,p}(\zeta)-f(\zeta))^2 ] + \frac{1}{P^2} \sum_{p=1}^{P} \sum_{p^{\prime}=1}^{P} \bbE [ (f^{N,p}(\zeta)-f(\zeta)) ] \bbE [(f^{N,p^{\prime}}(\zeta)-f(\zeta)) ]. \label{eqn:lem_eqn1}
\end{align}
By Proposition \ref{prop:single_smc_converge}, three expectation terms in \eqref{eqn:lem_eqn1} are expressed as follows
\begin{equation*}
    \bbE [ (f^{N,p}(\zeta)-f(\zeta))^2 ] \leq \frac{C|\zeta|_{\infty}^{2}}{N}, \ \bbE [ (f^{N,p}(\zeta)-f(\zeta)) ] = 0.
\end{equation*}
Then, we conclude
\begin{equation*}
    \bbE [(F^{N,P}(\zeta) - f(\zeta))^2] \leq \frac{C |\zeta|_{\infty}^{2}}{NP}.
\end{equation*}
\end{proof}

\subsection{Proof relating to Theorem \ref{thm:psmc_converge}}
\label{app:thm_psmc_converge}
{\em Given Assumptions \ref{ass:like_bound} and \ref{ass:mutation}, for suitable values of (\(M\),\(N\),\(J\))
    there exists a \(C_{\varphi} >0\), which depends on \(\varphi\), such that for any \(P \in \bbN\),
    \begin{equation} 
    \label{eqn:mse_pSMC}
    \bbE[(\hat{\varphi}_{\text{pSMC}} - \pi(\varphi))^2] \leq \frac{C_{\varphi}}{NP}.
    \end{equation}
    }
\begin{proof}
\begin{align}
   &\bbE[(\hat{\varphi}_{\text{pSMC}} - \pi(\varphi))^2] \\
   &= \bbE \bigg[ \bigg(\frac{F^{N,P}(\varphi)}{F^{N,P}(1)} - \frac{f(\varphi)}{f(1)} \bigg)^2 \bigg] \nonumber\\
    &= \bbE \bigg[ \bigg(\frac{F^{N,P}(\varphi)}{F^{N,P}(1)} - \frac{F^{N,P}(\varphi)}{f(1)} + \frac{F^{N,P}(\varphi)}{f(1)} - \frac{f(\varphi)}{f(1)} \bigg)^2 \bigg] \nonumber\\
    &\text{Applying Cauchy-Schwartz inequality, we have} \nonumber\\
    & \leq 2\bbE \bigg[ \bigg(\frac{F^{N,P}(\varphi)}{F^{N,P}(1)} - \frac{F^{N,P}(\varphi)}{f(1)} \bigg)^2 \bigg] + 2\bbE \bigg[ \bigg( \frac{F^{N,P}(\varphi)}{f(1)} - \frac{f(\varphi)}{f(1)} \bigg)^2 \bigg] \nonumber\\
    & \leq \frac{2|F^{N,P}(\varphi)|_{\infty}^2}{|F^{N,P}(1)|^2 |f(1)|^{2}} \bbE[(F^{N,P}(1)-f(1))^2] + \frac{2}{|f(1)|^2} \bbE \big[(F^{N,P}(\varphi)-f(\varphi))^2 \big] \label{eqn:thm_eqn1}
\end{align}
Assume Assumption \ref{ass:like_bound}, there exists a \(C_{\varphi}^{\prime}\) such that \(\frac{2|F^{N,P}(\varphi)|_{\infty}^2}{|F^{N,P}(1)|^2 |f(1)|^{2}} \leq C_{\varphi}^{\prime}\), and there exists a \(C^{\prime}\) such that \(\frac{2}{|f(1)|^2}  \leq C^{\prime}\). Then, following \eqref{eqn:thm_eqn1}, we have 
\begin{align}
    \bbE[(\hat{\varphi}_{\text{pSMC}} - \pi(\varphi))^2] &\leq  C_{\varphi}^{\prime} \bbE[(F^{N,P}(1)-f(1))^2] + C^{\prime} \bbE\big[(F^{N,P}(\varphi)-f(\varphi))^2 \big]  \nonumber\\
    & \text{By Lemma \ref{lem:psmc_single} with \(\zeta=\varphi\) and \(\zeta=1\) respectively, we have} \nonumber\\
    & \leq \frac{C_{\varphi}^{\prime} C}{NP} + \frac{C^{\prime} C |\varphi|_{\infty}^2}{NP} \nonumber\\
    & \text{Let } C_{\varphi}=C_{\varphi}^{\prime}C + C^{\prime} C |\varphi|_{\infty}^2 \text{, we have} \nonumber\\
    & \leq \frac{C_{\varphi}}{NP}. \nonumber
\end{align}
\end{proof}

\section{Complementary description of simulations}
\label{app:simos}

\subsection{Computation of Error bars}
\label{app:errorbar}

Assume running $R$ times of experiments to get $R$ square errors/loss between simulated estimator $\hat{\varphi}$ and the ground truth, SE$(\hat{\varphi})^{r}$ for $r=1,...,R$. Take the MSE as an example, the MSE is the mean of SE$(\hat{\varphi})^{r}$ over $R$ realizations, and the standard error of MSE (s.e.) is computed by
\begin{equation}
    \frac{\sqrt{\frac{1}{R}\sum_{r=1}^{R}(\text{SE}(\hat{\varphi})^{r}-\text{MSE})^2}}{\sqrt{R}}.
\end{equation} 

\subsection{Integrated Autocorrelation Time}
\label{app:IACT}

Integrated Autocorrelation Time (IACT) means 
the time until the chain is uncorrelated with its initial condition.
The precise mathematical definition is as follows.

Let $\theta_0, \dots, \theta_t, \dots$ denote the Markov chain, and let $\varphi(\theta)$ be a scalar function of the state. 
We first define the \emph{autocovariance function} (ACF) at lag $s$:
\[
\gamma_s(\varphi) 
= \mathbb{E}\!\left[ \big(\varphi(\theta_{t+s}) - \mathbb{E}[\varphi(\theta)]\big)
                     \big(\varphi(\theta_t) - \mathbb{E}[\varphi(\theta)]\big) \right],
\]
and the ACF at lag $s$ as the normalized quantity
\[
\rho_s(\varphi) = \frac{\gamma_s(\varphi)}{\gamma_0(\varphi)},
\]
where $\gamma_0(\varphi)$ is the variance of $\varphi(\theta)$.

Then the \emph{integrated autocorrelation time} (IACT) of $\varphi$ is then defined in terms of the ACF by
\[
\text{IACT}(\varphi) = 1 + 2 \sum_{s=1}^{\infty} \rho_s(\varphi).
\]

\subsection{Details of Gaussian cases}
\label{app:Gaussian}

Assume we have \(y \in \bbR^{m}\), \(X \in \bbR^{m\times d}\) and parameter \(\theta \in \bbR^{d}\) connected by the following inverse problem:
\begin{equation}
    y=X\theta + \nu, \ \nu \sim \mathcal{N}(0,\sigma^2I_m),
\end{equation}
where \(X\) is the design matrix. If we let \(\pi_0(\theta) = \mathcal{N}(\mu_0,\Sigma_0)\), this is one of the very few problems with an analytical Bayesian posterior, which will provide a convenient ground truth for measuring convergence. In particular, the posterior distribution is a multivariate Gaussian distribution \(\mathcal{N}(\mu,\Sigma)\), where
\[
\mu = \Sigma(\Sigma_0^{-1} \mu_0 + \frac{1}{\sigma^2} X^Ty), \ \Sigma = (\Sigma_0^{-1} + \frac{1}{\sigma^2} X^T X)^{-1}.
\]
See e.g. \cite{bishop2006pattern}.

Let \(X \in \bbR^{m \times d}\) be a randomly selected full rank matrix and \(\sigma = 0.01\). The observations are generated as
\begin{equation}
    y = X\theta^* + \nu,
\end{equation}
where \(\theta^*\sim \pi_0\) and \(\nu\sim \mathcal{N}(0,\sigma^2 I_m)\) are independent.

\subsection{Details of the Bayesian Neural Networks}
\label{app:BNN}

Let weights be \(A_{i} \in \bbR^{n_i \times n_{i-1}}\) and biases be \(b_i \in \bbR^{n_i}\) for \(i \in \{1,...,D\}\), we denote 
\(\theta := ((A_1,b_1),...,(A_D,b_D))\). The layer is defined by
\begin{equation*}
\begin{aligned}
    g_1(x,\theta) & := A_1 x + b_1, \\
    g_d(x,\theta) & := A_i \sigma_{n_{i-1}}(g_{i-1}(x)) + b_{i}, \ \ i \in \{2,...,D-1\}, \\
    g(x,\theta) & := A_{D}\sigma_{n_{D-1}}(g_{D-1}(x)) + b_{D},
\end{aligned}
\end{equation*}
where \(\sigma_{i}(u) := (\nu(u_1),...,\nu(u_i))^{T}\) with ReLU activation \(\nu(u) = \max \{0,u\}\).

Consider the discrete data set in a classification problem, we have \(\sY = \{1,...,K\}\) and \(n_D=K\), then we instead define the so-called \textit{softmax} function as
\begin{equation}\label{eqn:bayes_class}
    h_k(x,\theta) = \frac{\exp(g_{k}(x,\theta))}{\sum_{j=1}^{K}\exp(g_{j}(x,\theta))}, \ k \in \sY,
\end{equation}
and define \(h(x,\theta) = (h_1(x,\theta),...,h_K(x,\theta))\) as a categorical distribution on \(K\) outcomes based on data \(x\). Then we assume that \(y_i \sim h(x_i)\) for \(i=\{1,...,m\}\).



Now we describe the various neural network architectures we use for the various datasets.

\subsubsection{MNIST Classification Example}
\label{app:mnist}

The architecture is a simple CNN with 
(i) one hidden layer with 
$4$ channels of $3\times3$ kernels with unit stride and padding,
followed by (ii) ReLU activation and 
(iii) $2\times2$ max pooling, (iv) a linear layer,
and (v) a softmax. 
The parameter prior and dataset for MNIST are as follows. The parameter prior is independent $\mathcal{N}(0,\sigma^2)$, with $\sigma^2$ on weights and biases determined by Kaiming setting. Here we consider a subset of $1000$ train and $1000$ test data, both as a warm-up and also because there is very little uncertainty remaining for models trained with the full dataset, which makes it less
    interesting for BDL.

\subsubsection{IMDb Classification Example}
\label{app:imdb}

Here we use SBERT embeddings \cite{reimers-2019-sentence-bert} based on the model 
{\texttt all-mpnet-base-v2} \cite{song2020mpnet}
\footnote{https://huggingface.co/sentence-transformers/all-mpnet-base-v2}.
In other words, frozen weights from {\texttt all-mpnet-base-v2}
until the $768$ dimensional [CLS] output.
The NN model and parameter prior for IMDb\footnote{https://huggingface.co/datasets/stanfordnlp/imdb} experiment are as follows.
NN is followed by (i) one hidden layer with $128$ neurons, (ii) ReLU activation, (iii) a final linear layer, and (iv) softmax output. The parameter prior is independent $\mathcal{N}(0,\sigma^2{\sf Id})$, where \(d=98690\), with $\sigma=0.1$ on weights and $\sigma=0.01$ on biases. The whole train ($25000$ data) and test dataset ($25000$ data) are considered.

\subsubsection{CIFAR-10 Classification Example}
\label{app:cifar}

Here, the architecture is ResNet-50 pre-trained from ImageNet with 
all parameters frozen until the final pooled $2048$ dimensional features.
The NN model and parameter prior for CIFAR10 experiments are as follows. NN is followed by (i) one hidden layer with $128$ neurons, (ii) ReLU activations, (iii) a final linear layer, and (iv) softmax output. The parameter prior is independent $\mathcal{N}(0,\sigma^2 {\sf Id})$, where \(d=263562\), with $\sigma=0.01$ on weights and $\sigma=0.001$ on biases. The whole train ($50000$ data) and test dataset ($10000$ data) are considered.

\subsection{Hardware description}

The main CPU cluster we use  
has nodes with 2 $\times$ 16-core Intel Skylake Gold 6130 CPU @ 2.10GHz, 192GB RAM
{\em without communication} in between,
so it can only run $N/P=32$ particles in parallel
with one particle per core. 
There are also unconnected AMD 
“Genoa” compute nodes, 
with 2 $\times$ 84-core AMD EPYC 9634 CPUs and  
1.5TB RAM. 
There is a pool cluster with 
4096 cores in total provided by 128 $\times$ Skylake nodes,
connected by 
Mellanox Technologies MT27800 Family [ConnectX-5] 100Gb/s InfiniBand interconnect,
which requires special permissions.
This is used only for the inter-connect experiments presented in Figure \ref{fig:mnist_hpc_inter_connect}.

\section{Further results and description}

\subsection{Free parameters in SMC and pCN}
\label{app:free}

Experiments in this section are tested on the Gaussian case defined in Appendix \ref{app:Gaussian}.

MCMC has some free parameters that need to be selected. 
For both pCN and (fixed trajectory) HMC, the minimal tuning parameter is the step size,
and we constrain our attention to this, to avoid unnecessary complication.
The mixing time $T$ will be minimized if the step size is chosen optimally 
to target a suitable acceptance probability \cite{beskos2013optimal,cotter2013mcmc},
and this is achieved adaptively.
The number of warm-up samples $B$ needs to be chosen, and should be $bT$, as discussed above.

Note that other more sophisticated tuning, such as adapting the mass matrix 
can potentially improve the mixing of MCMC \cite{carpenter2017stan}, 
both on its own and within SMC \cite{buchholz2021adaptive}.

If $J$ is selected as described in Example \ref{ex:adaptive_temp}, then the remaining 
free parameters for SMC beyond those of the MCMC kernel are $(M,N)$. 
Numerical experiments suggest that the relationship for $M\geq M_{\rm min}, N\geq N_{\rm min}$
is approximately $C/NM^a$, for $a\leq 1/2$ and decreasing with problem complexity or mixing time.
Therefore, we aim to let $M=M_{\rm min}$; however, it is not clear how to select $M_{\rm min}$.
In practice, we have found similar results by choosing $M$ adaptively. These results 
are given in the Appendix \ref{sec:varyNM}.

\subsubsection{How to select a suitable \(N\) and \(M\) in a single SMC}
\label{sec:varyNM}

We conduct experiments by varying \( N \in [1, 2^{14}] \) and \( M \in [1, \tau_{\varphi} = 72160] \). Fixing \( N \) and varying \( M \) yields Figure~\ref{fig:fixN_varyM}, while fixing \( M \) and varying \( N_{\text{SMC}} \) produces Figure~\ref{fig:fixM_varyN}.

\begin{figure}[H]
    \centering
    \begin{subfigure}[b]{0.49\textwidth}
    \centering
        \includegraphics[width=\columnwidth]{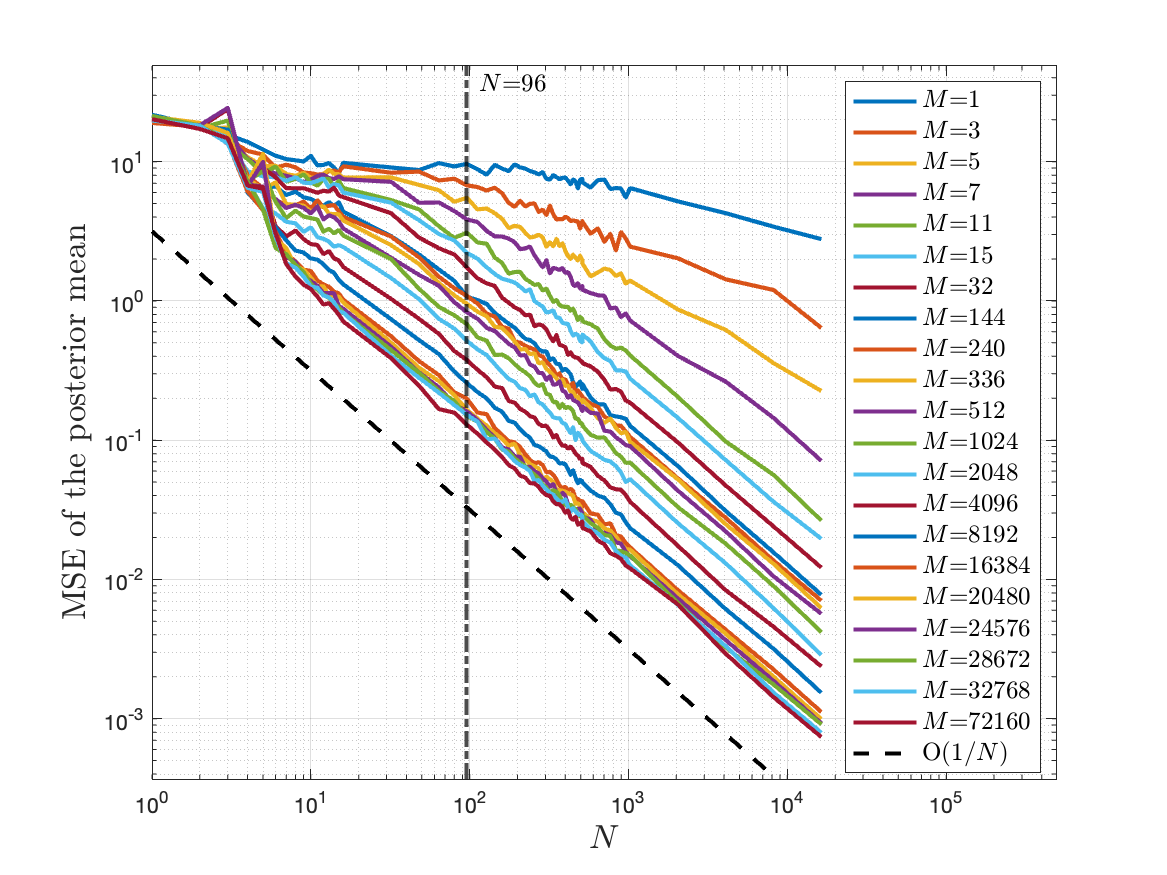}
        \caption{Fix \( M \) and vary \( N \)}
        \label{fig:fixM_varyN}
    \end{subfigure}
    \begin{subfigure}[b]{0.49\textwidth}
    \centering
        \includegraphics[width=\columnwidth]{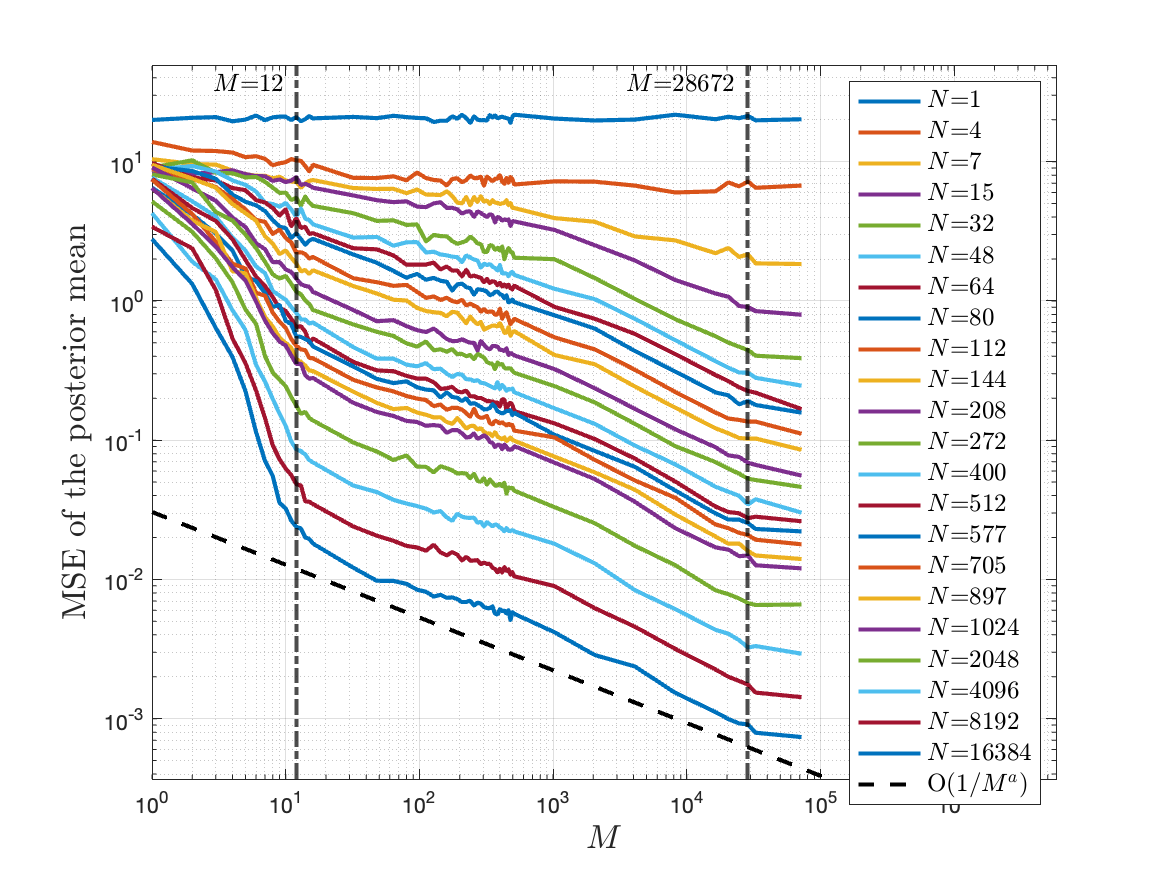}
        \caption{Fix \( N \) and vary \( M \)}
        \label{fig:fixN_varyM}
    \end{subfigure}
    \caption{MSE of the posterior mean as a function of \( N \) and \( M \), on Gaussian case.}
\label{fig:changeNM}
\end{figure}

The key conclusion is that both \( M \geq M_{\sf min} \) and \( N \geq N_{\sf min} \) are necessary to achieve ``regular behavior'' in the performance of the SMC estimator, i.e. the \(-1\) convergence rate. When \( M \) is too small, the constants involved in the error bounds become prohibitively large, even though the estimator remains consistent. On the other hand, for very large \( M \), the marginal benefit in terms of estimator accuracy diminishes rapidly, implying an unfavorable exchange rate between increased \( M \) and the computational cost. Moreover, \( M_{\sf min} \) is not fixed—it depends on \( N \); smaller values of \( N \) generally require larger values of \( M \) to compensate.

Figure~\ref{fig:fixM_varyN} reveals a sharp transition in behavior around \( M \approx 16 \): for \( M < 16 \), increasing \( N \) yields a noticeable improvement in performance, following roughly a \( 1/M \) rate; beyond this point, improvements become flatter. Additionally, we observe that increasing \( M \) beyond \( 28672 \approx 0.4\tau_{\varphi} \) results in negligible further reduction in the MSE.

A more detailed analysis of Figure~\ref{fig:fixN_varyM} for sufficiently large \( N \geq 96 \) reveals three distinct regimes for the dependence of MSE on \( M \):
\begin{itemize}
    \item For \( M < 12 \), MSE decreases rapidly and nonlinearly on a logarithmic scale. The rate of decay becomes more significant as \( N_{\text{SMC}} \) increases.
    \item For \( 12 \leq M \leq 28672 \), the MSE exhibits an approximately linear decline in \( \log \)-scale with respect to \( M \).
    \item For \( M > 28672 \), MSE plateauing occurs, with minimal gains from further increasing \( M \).
\end{itemize}
To better understand the linear decrease pattern, we perform logarithmic linear regression to estimate the parameters 
\(a\) and \(C\) in the empirical model:
\begin{equation*}
    \log(\text{MSE})=C - \log(N) - a\log(M).
\end{equation*}
We define
\begin{align*}
    &Y=\log(\text{MSE})+\log(N)\\
    &X=\log(M)
\end{align*}
and it becomes \(Y = C-aX\). Then, we apply linear regression to estimate the parameters, obtaining the fitted values \(\hat{a}=0.3789\) and \(\hat{C}=829.5622\).To evaluate the model's accuracy, we compute the MSE and Mean Absolute Percentage Error (MAPE) between the predicted MSE values (\(\hat{\text{MSE}}\)) and the true MSE values (\(\text{MSE}\)). Here, MSE refers to the mean squared error of the posterior mean estimated by SMC for different values of \(N\) and \(M\). Let \(n_1\) be the number of values of \(N\) used and \(n_2\) be the number of values of \(M\) used, then \(\hat{\text{MSE}}_{i,j}\) and MSE\(_{i,j}\) represent the predicted and true MSE values, respectively, for different values of \(N\) and \(M\), where \(i=1,...n_1\) and \(j=1,...n_2\).

\begin{itemize}
    \item MSE of the predicated data and the true data is computed by
    \begin{equation*}
        0.006939 = \frac{1}{n_1n_2}\sum_{i,j} (\hat{\text{MSE}}_{i,j}-\text{MSE}_{i,j})^2 
    \end{equation*}
    \item MAPE of the predicated data and the true data is computed by
    \begin{equation*}
        9.30 = \frac{100}{n_1n_2}\sum_{i,j} \Bigg|\frac{\hat{\text{MSE}}_{i,j}-\text{MSE}_{i,j}}{\text{MSE}_{i,j}}\Bigg| 
    \end{equation*}
\end{itemize}

The low MSE suggests that the logarithmic linear regression model fits the data well, with small absolute deviations. Moreover, the MAPE is 9.30\% means that, on average, the predicted MSE values deviate by about 9.30\% from the true MSE values. This is a relatively small percentage error, suggesting that the model provides reasonably accurate predictions.


These observations provide practical guidelines for selecting suitable values of \( M \) and \( N \) in SMC experiments: when \( N \geq 100 \) and \( M \geq 16 \), the convergence rate stabilizes at approximately \(-1\), suggesting that this regime avoids significant bias and variance. This guides the selection of the minimal value of \(N\) and \(M\) for the following experiments.


\subsubsection{Analysis on Adaptive SMC}


From Figure \ref{fig:MSE_adaptive_vs_fixed}, we observe that for a reasonable \( N = 200 \), the performance of SMC and SMC with adaptive \( M \) remains similar when considering the same computational cost. Additionally, Figure~\ref{fig:MSE_adaptive_vs_fixed} indicates that the transition point for \( \eta \) occurs at \( \eta = 0.05 \). This suggests that the minimum practical value for \( \eta \) is \(0.05\), where the transition point aligns closely with that of SMC using a fixed \( M \).

For small $N$, SMC with adaptive $M$ tends to perform similarly to a carefully selected fixed $M$, offering a viable alternative when the optimal fixed $M$ is unknown. However, to consistently observe regular behavior in autocorrelation metrics and to fully benefit from adaptive $M$, a relatively large $N$ is required. We observed that small sample sizes make adaptation challenging due to the high variance inherent in such statistics. As $N$ increases, the adaptive mechanism stabilizes and behaves more predictably, aligning with the expected geometric growth from $M_0$ to $M_J$. This behavior is supported by the results shown in Figure~\ref{fig:adaptive_M_scaling}.

\begin{figure}[H]
    \centering
    \begin{subfigure}[b]{0.49\textwidth}
    \centering
        \includegraphics[width=\columnwidth]{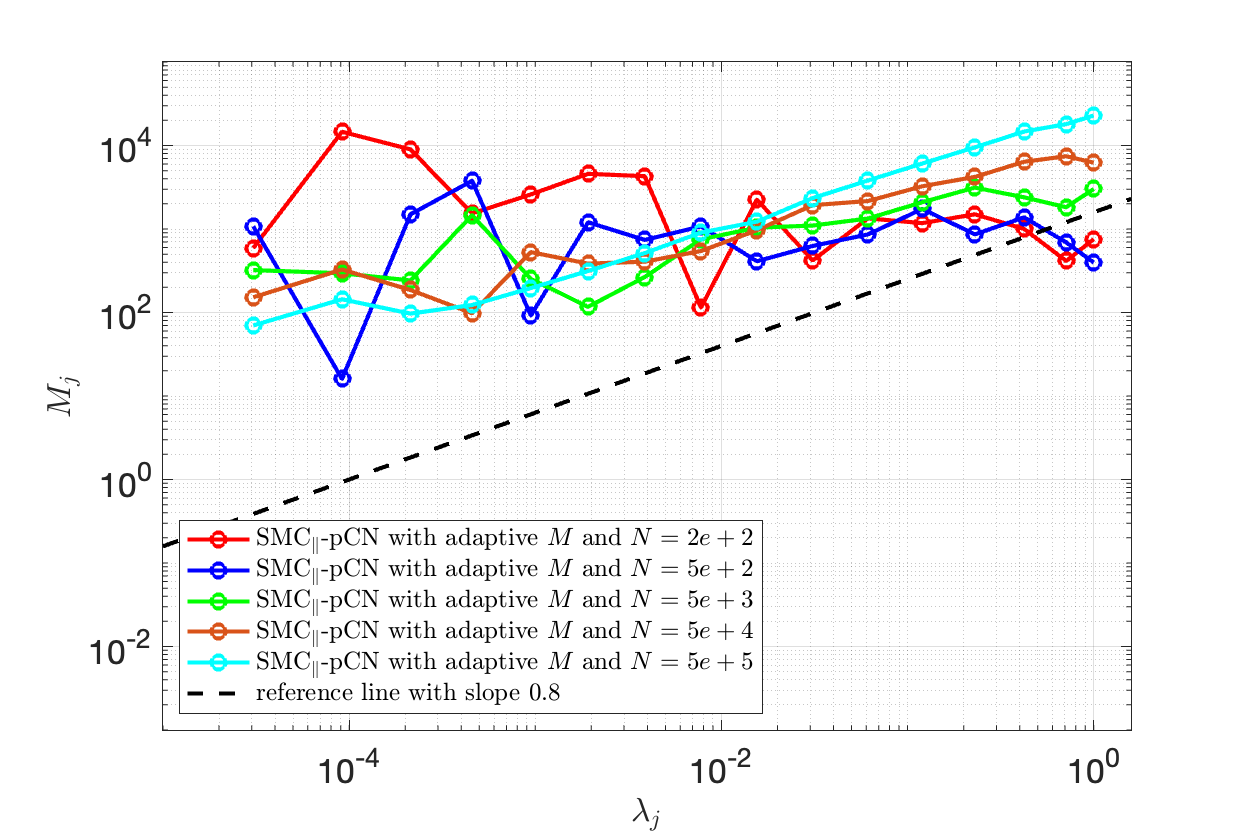}
        \caption{}
        \label{fig:adaptive_M_scaling}
    \end{subfigure}
    \begin{subfigure}[b]{0.49\textwidth}
    \centering
        \includegraphics[width=\columnwidth]{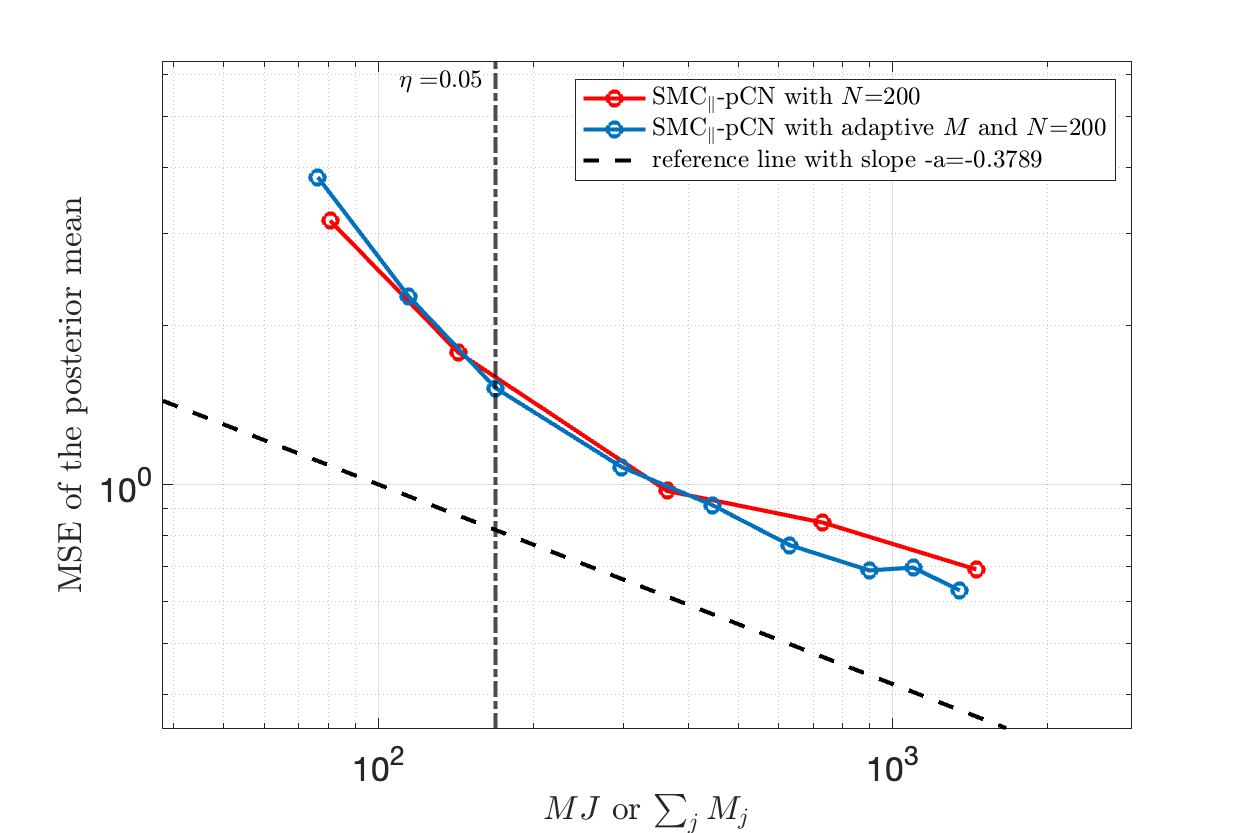}
        \caption{}
        \label{fig:MSE_adaptive_vs_fixed}
    \end{subfigure}
    \caption{Analysis of SMC with fixed and adaptive \( M \), on Gaussian case. (a) Scaling of \( M_{j}\) with different values of \( N \) in adaptive SMC. (b) MSE of SMC with fixed and adaptive \( M \).}
\end{figure}

\subsubsection{Cost ablation for Gaussian case}
\label{sec:ffc-gauss}

\begin{wraptable}[9]{r}{0.55\textwidth}
    \vspace{-5pt}
    \centering
    \caption{Relative MSE ($\downarrow$) for SMC and pCN at convergence and away from convergence, on Gaussian case.
    }
    \begin{tabular}{c|c|c|c}
    \toprule
    & High Cost & Medium Cost & Low Cost\\
    \midrule
    SMC ($\downarrow$) & 0.0563 & 0.1237 & 0.2160 \\
    \midrule
    pCN ($\downarrow$) & 0.0676 & 0.3663 & 1.7411 \\
    \bottomrule
    \end{tabular}
    \label{tab:SMC_vs_pCN_samecost}
\end{wraptable}
Table \ref{tab:SMC_vs_pCN_samecost} presents relative MSE ($\downarrow$)  
for converged and far-from-converged pCN MCMC and SMC with pCN 
applied to a tractable Gaussian example with 4 high-accuracy observations
of a $d=16$ dimensional parameter. This is a difficult regime for pCN.
    SMC always takes $M=T/2J$ pCN mutation steps with $N=200$ particles, 
    while pCN takes $TN$ steps with $N=100$ (equivalent costs).
    Three cost regimes are considered: 
    high (\(T=T_{A}\))
    medium (\(T=T_{A}/10\)), 
    and low (\(T=T_{A}/100\)). 
This shows that 
(i) the methods perform comparably in terms of total computational cost and
(ii) SMC performs better far from convergence.
HMC is more robust, and we have seen in Figure \ref{fig:singletons}
stating that (ii) is not generic.

\subsection{Communication overhead}
\label{app:interconnect}

Experiments in this section are tested on the MNIST dataset with the model setting stated in Appendix \ref{app:mnist}.

\paragraph{Interconnect: HPC Pool Performance between SMC and SMC\(_\parallel\).}
Figure \ref{fig:mnist_hpc_inter_connect} (also see Figure \ref{fig:mnist_metric_fixtraj-main} in paper)
presents the performance of a SMC algorithm with \( NP \) particles executed across \( P \) nodes on the HPC pool, compared to parallel SMC (PSMC), where each of the \( P \) instances of SMC with \( N \) particles runs independently on separate nodes. In this configuration, SMC\(_\parallel\) is executed on a single Skylake node, while the SMC utilizes \( P \) nodes simultaneously on an HPC system, with each node providing 32 cores. The communication overhead inherent in the single SMC setup becomes evident: for \( N = 128 \), the execution time increases by approximately 40\%, and for \( N = 256 \), by 72\%, relative to the parallel SMC with \( N = 32 \). This implies an approximate time cost increase of \( 40\log_2(P/2) \)\% when scaling up to \( N = 32P \) in the single SMC setting. In Figure \ref{fig:mnist_hpc_inter_connect}, results for $N=16$ and $N=8$ are included, indicating that SMC$_\parallel$ breaks when $N$ is too small. The full data result is given in Table \ref{tab:mnist_hpc_inter_connect}.

\begin{figure}[H]
    \centering
    \includegraphics[width=\columnwidth]{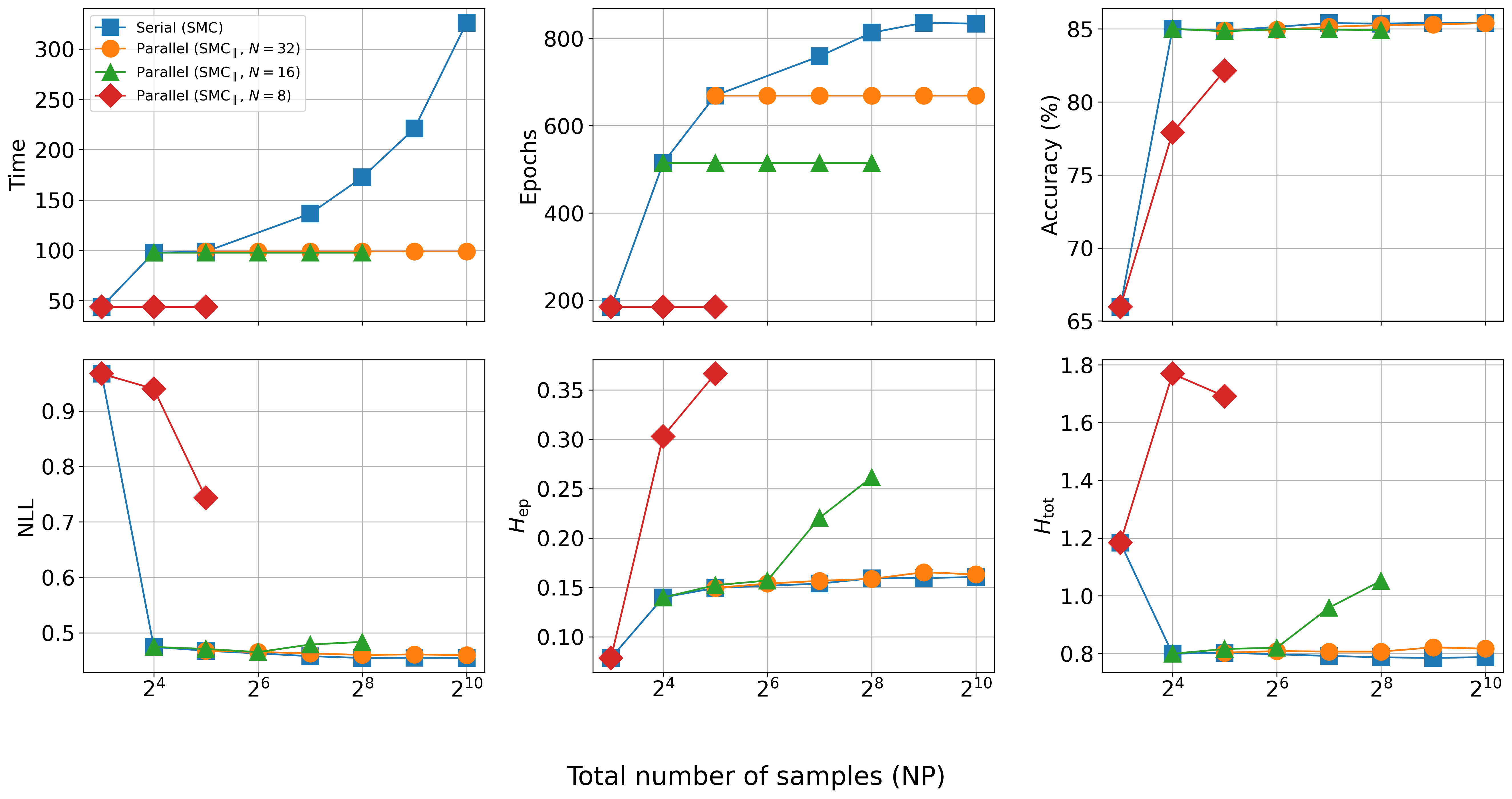}
    \caption{Comparison of SMC\(_\parallel\) (\(P\) chains with \(N\) particles) and SMC (\(NP\) samples), with fixed trajectory \(L\delta=0.1\) and \(M=1\).} 
    \label{fig:mnist_hpc_inter_connect}
\end{figure}

\paragraph{Parallel scaling in SMC and single serial HMC running in nodes with different number of cores.} Table~\ref{tab:mnist_hpc_differcore} presents the performance results of the SMC and HMC methods run on nodes with different numbers of cores. 
This is acceleration from single-node multi-core parallel multi-threading of serial processes.
A clear reduction in computation time can be observed as the setup scales from a single core to a 32-core configuration, where the number of cores matches the number of samples used. In the case of SMC, parallelization is achieved by assigning each sample to a separate core, allowing efficient computation with inter-process communication occurring as needed, for example, during resampling or adaptive tempering. In contrast, the HMC method leverages parallelism internally through libraries such as PyTorch and NumPy.

\begin{table}[H]
\scriptsize
\caption{Comparison of SMC (\(N=32\) particles) 
and {single HMC chain (
approximately 32 independent samples),
with fixed trajectory length $\tau=0.1$,}
\(B=T=900\) and \(M=20\) (\(10\) realizations and \(\pm\) s.e. in accuracy).}
\label{tab:mnist_hpc_differcore}
\centering
\begin{tabular}{l|cccc|ccccc}
\toprule
& \multicolumn{4}{c|}{\(1\) core} & \multicolumn{5}{c}{\(32\) core} \\
\cmidrule(lr){2-5} \cmidrule(lr){6-10}
& Acc. & Time & Epochs. & Epochs/s & Acc. & Time & Speedup & Epochs. & Epochs./s \\
\midrule
HMC & 85.40\(\pm\)0.11 & 8.611e+04  & 8.954e+05 & 10.4 & 85.26\(\pm\)0.12 
& 1.103e+04 & 7.8 & 9.352e+05 & 84.8 \\
\midrule
SMC & 85.53\(\pm\)0.11 & 5.840e+04 & 6.421e+05 & 11.0 & 85.33\(\pm\)0.07 
& 1.825e+03 & 32.0 & 1.561e+04 & 
8.5($\times 32$) \\
\bottomrule
\end{tabular}
\end{table}

\subsection{Exchangeability between \(N\) and \(P\)}
\label{app:exchange}

\paragraph{Gaussian example.} Experiment in Figure \ref{fig:trade_NP} is tested on the Gaussian case defined in Appendix \ref{app:Gaussian}. We analyse how the MSE of the posterior mean for different values of \( M \), under fixed \( NP = 1024 \), in Figure~\ref{fig:trade_NP}. We highlight two important findings:
\begin{itemize}
    \item The performance stabilizes for \( M \geq 16 \), beyond which increasing \( M \) yields diminishing improvements in MSE. This supports the identification of \( M = 16 \) as a practical lower bound for ensuring stable estimation.
    \item When \( N \geq 32 \), the estimator shows a form of \emph{work equivalence} between \( N \) and \( P \), meaning that increasing \( N \) while reducing \( P \) (or vice versa) leads to similar MSE behavior. This suggests that for proper \( N \), one can trade off between the number of particles and the number of parallel chains without a sacrifice in accuracy.
\end{itemize}

\begin{figure}[H]
    \centering
    \includegraphics[width=.6\columnwidth]{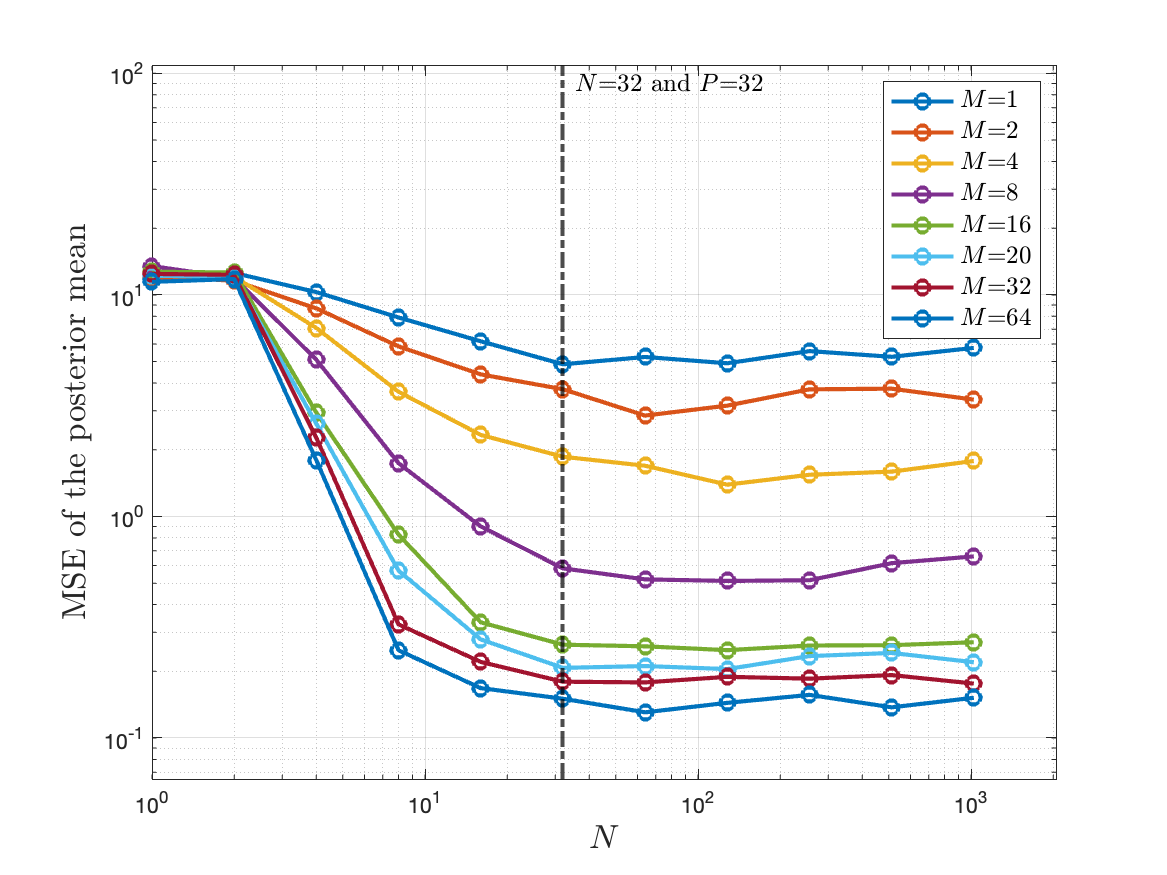}
    \caption{MSE of the posterior mean for different \( M \), with fixed \( NP = 1024 \), on Gaussian case (100 realizations).}
    \label{fig:trade_NP}
\end{figure}

\paragraph{MNIST example.} Experiments in Table \ref{tab:mnist_metric_fixtraj0005_diffN} and \ref{tab:mnist_metric_fixtraj002_diffN} are tested on the MNIST dataset with the model setting stated in Appendix \ref{app:mnist}. We analyse how the accuracy and NLL are under fixed $NP$ with a sufficiently large $M$ chosen. We observed that the accuracy and NLL reach a similar performance for $N=32$ and $N=64$, which are better than the results for $N=16$ and $N=8$, suggesting that $N=32$ is also a proper choice for MNIST experiments.

\begin{table}[H]
\footnotesize
\caption{Comparison of different settings of SMC\(_\parallel\) (\(P\) SMC and \(N\) particles), with fixed trajectory \(L\delta=0.005\) and \(M=20\), on MNIST (\(10\) realizations and \(\pm\) s.e. in accuracy).}
\label{tab:mnist_metric_fixtraj0005_diffN}
\centering
\begin{tabular}{l|cccc|cccc}
\toprule
\(NP\) & \multicolumn{4}{c|}{Acc.} & \multicolumn{4}{c}{NLL} \\
\cmidrule{2-5} \cmidrule{6-9}
& \(N=64\) & \(N=32\)& \(N=16\) & \(N=8\)
& \(N=64\) & \(N=32\) & \(N=16\) & \(N=8\) \\
\midrule
\(32\)  & – & 82.12\(\pm\)0.19 & 81.63\(\pm\)0.16 & 81.47\(\pm\)0.09
        & 6.166e-1 & 6.270e-1 & 6.552e-1 & – \\
\midrule
\(64\)  & 82.39\(\pm\)0.26 & 82.69\(\pm\)0.10 & 81.91\(\pm\)0.10 & 81.69\(\pm\)0.05
        & 5.966e-1 & 6.010e-1 & 6.209e-1 & 6.470e-1 \\
\midrule
\(128\) & 82.85\(\pm\)0.11 & 82.98\(\pm\)0.16 & 82.64\(\pm\)0.05 & 81.97\(\pm\)0.15
        & 5.847e-1 & 5.950e-1 & 6.144e-1 & 6.463e-1 \\
\bottomrule
\end{tabular}
\end{table}

\begin{table}[H]
\footnotesize
\caption{Comparison of different settings of SMC\(_\parallel\) (\(P\) SMC and \(N\) particles), with fixed trajectory \(L\delta=0.02\) and \(M=20\), on MNIST (\(10\) realizations and \(\pm\) s.e. in accuracy).}
\label{tab:mnist_metric_fixtraj002_diffN}
\centering
\begin{tabular}{l|cccc|cccc}
\toprule
\(NP\) & \multicolumn{4}{c|}{Acc.} & \multicolumn{4}{c}{NLL} \\
\cmidrule{2-5} \cmidrule{6-9}
& \(N=64\) & \(N=32\) & \(N=16\) & \(N=8\)
& \(N=64\) & \(N=32\) & \(N=16\) & \(N=8\) \\
\midrule
\(32\)  & -                & 84.93\(\pm\)0.08 & 84.63\(\pm\)0.04 & 84.51\(\pm\)0.03
        & -                & 4.800e-1         & 4.918e-1         & 5.127e-1         \\
\midrule
\(64\)  & 85.11\(\pm\)0.06 & 84.83\(\pm\)0.04 & 84.85\(\pm\)0.03 & 84.53\(\pm\)0.04
        & 4.679e-1         & 4.758e-1         & 4.886e-1         & 5.110e-1         \\
\midrule
\(128\) & 85.22\(\pm\)0.04 & 84.98\(\pm\)0.01 & 84.92\(\pm\)0.01 & 84.68\(\pm\)0.02
        & 4.692e-1         & 4.756e-1         & 4.851e-1         & 5.091e-1         \\
\bottomrule
\end{tabular}
\end{table}

These results also align with the expected \( 1/N \) convergence behavior for single-chain estimators, which supports that this exchangeability between \(N\) and \(P\) holds under broader conditions.

\subsection{Large $P$ trade-off between SMC and SMC$_\parallel$}
\label{app:par_gauss}

Experiments in this section are tested on the Gaussian case defined in Appendix \ref{app:Gaussian}.

From the investigation in Section \ref{app:exchange}
, we let \(M=16\) to make sure the MCMC kernel is well-mixed. 
Let the 
The total number of samples is \(2^{16}\), and results correspond to \(P=2^{0:1:14}\) 
single SMCs with \(N=2^{16}/P\) particles. 
The relative difference of MSE between single SMC and SMC$_\parallel$ is plotted
in Figure \ref{fig:SMC_vs_PSMC}.
Note that the single monolithic SMC is \(N\)-parallel {\em with communication}, 
whereas the $P$-parallelism 
is {\em free from any communication}.
Assuming that a relative difference in MSE above $0.5$ is negligible, 
this provides a limit for asynchronous parallelism, which is $N=16$ in the plot.
There is limited loss down to $16$ cores, which
is convenient since modern SIMD nodes
typically have at least $16$ cores 
and intra-node communication 
is usually very efficient,
as opposed to between-node inter-connect.






\begin{figure}[H]
    \centering
    \includegraphics[width=.6\columnwidth]{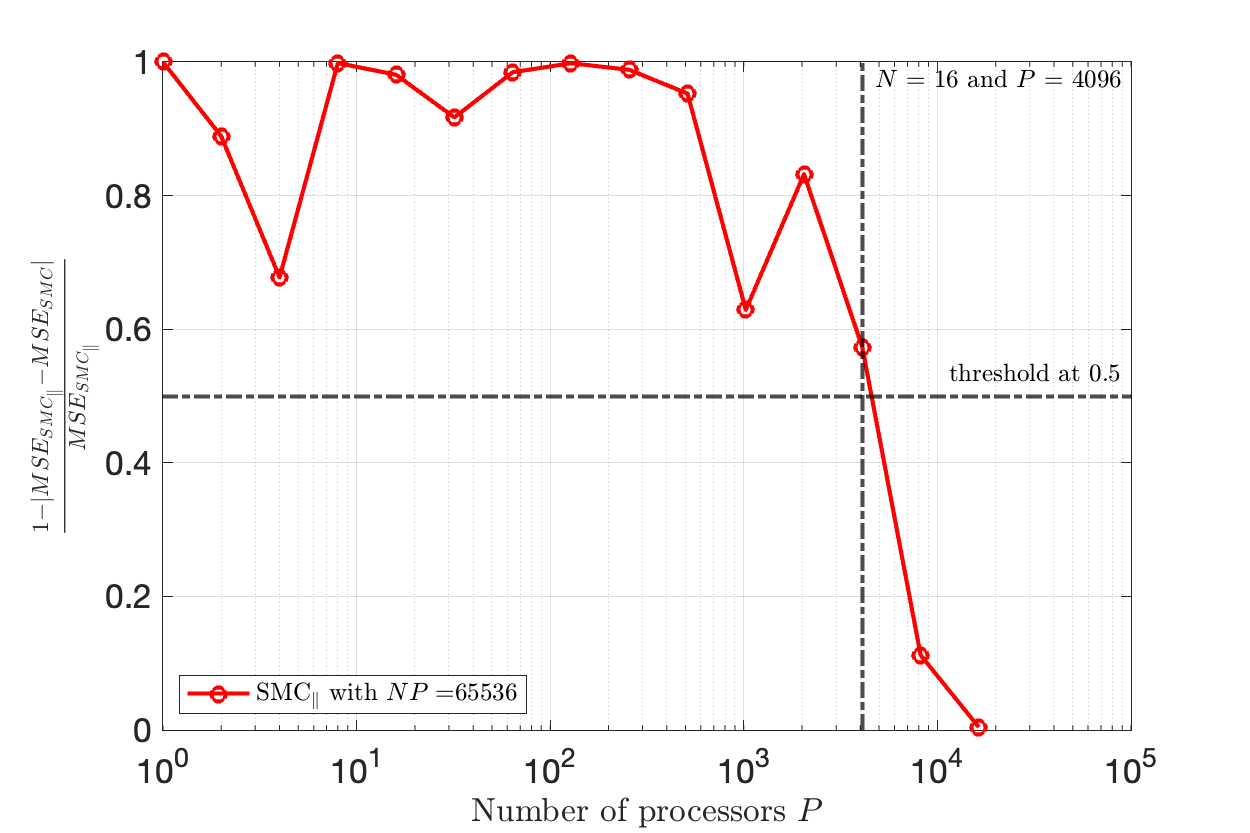}
    \caption{SMC$_\parallel$ with 
    \(NP=2^{16}\) samples, 
    where individual SMCs have $N$ particles and
    \(P\) increases.
    } 
\label{fig:SMC_vs_PSMC}
\end{figure}

\subsection{Full comparison of SMC$_\parallel$ and HMC$_\parallel$ on MNIST}
\label{app:mnist_full}

Experiments in this section are tested on the MNIST dataset with the model setting stated in Appendix \ref{app:mnist}.

\textbf{Different fixed trajectories ($\tau$)} Comparison among SMC\(_\parallel\), HMC\(_\parallel\), HMC and HMC (GS) with different $\tau$ and increasing \(P\) is given in Figure \ref{fig:mnist_metric_fixtraj-main} in the main text, where the gold-standard is computed by HMC (HMC (GS)) over 5 realizations with \(N=B=1e5\) (\(T=1\)) for different $\tau$. The full data results for Figure \ref{fig:mnist_metric_fixtraj-main} for \(\tau=0.005, 0.02, 0.1\) are given in the Table \ref{tab:mnist_metric_fixtraj0005}
, \ref{tab:mnist_metric_fixtraj002} 
and \ref{tab:mnist_metric_fixtraj01}.

\begin{wraptable}[16]{r}{0.5\textwidth}
\setlength{\tabcolsep}{3pt}
\vspace{-10pt}
\footnotesize
\caption{Comparison of single process and {\em intra-parallel} SMC and HMC for different $\tau$.}
\label{tab:time_node}
\centering
\begin{tabular}{l|cc|cc|cc}
\toprule
& \multicolumn{2}{c|}{\textbf{\(\tau=0.005\)}} & \multicolumn{2}{c|}{\textbf{\(\tau=0.02\)}} & \multicolumn{2}{c}{\textbf{\(\tau=0.1\)}} \\
serial & SMC & HMC & SMC & HMC  & SMC & HMC \\
\midrule
Acc.($\uparrow$) & 82.12 & 84.92 & 84.93 & 85.17 & 85.53 & 85.35\\
\midrule
NLL ($\downarrow$) & 616.6 & 475.7 & 480.0 & 453.3 & 450.9 & 454.9\\
\midrule
epochs & 2.9e4 & 3.0e4 & 6.5e4 & 1.8e5 & 5.3e5 & 9.4e5 \\
\midrule
intra-$\parallel$\\
\midrule
Acc.($\uparrow$) & 82.12 & 83.16 & 84.93 & 85.30 & 85.53 & 85.20\\
\midrule
NLL ($\downarrow$) & 616.6 & 596.1 & 480.0 & 455.8 & 450.9 & 452.9\\
\midrule
epochs
& 906 & 945 & 2040 & 5512 & 1.6e4 & 2.9e4\\
\bottomrule
\end{tabular}
\end{wraptable}
We now take a closer look at some selected data in these results and regenerate it as Table \ref{tab:time_node}. This table present results with fixed trajectory lengths of $\tau\in\{0.005, 0.02, 0.1\}$ and step size adapted for $0.65$ acceptance rate. SMC uses \(M=20\) HMC steps per adaptive tempering step with \(N=32\) particles, while HMC runs for \(NT\) steps with $T=900$. All results average 10 realizations. The serial versions use a single core. {\em Intra-parallel} SMC leverages intrinsic parallelism across $N$ cores and parallel HMC runs $N$ separate chains for $B=T$ steps only. This shows that (i) SMC and HMC achieve comparable results when run for long enough  
(around $1e3-1e4$ epochs here), (ii) intra-parallel SMC achieves identical results and parallel HMC achieves similar results for longer trajectories, {\em with $N-$fold speedup},
quantified by number of epochs, and (iii) for short trajectory, SMC and parallel HMC perform worse than serial HMC.

\paragraph{Low cost regime (\(L=1\)).} Figure~\ref{fig:mnist_metric_diffM} illustrates an extreme case with $L=1$,
where the model converges to a sub-optimal plateau, as we observed in Figure~\ref {fig:mnist_metric_diffM} for $\tau=0.005$ in the main text. Notably, convergence often occurs at the very first point ($P=1$) across most settings. While increasing $M$ leads to gradual improvement of the sub-optimal plateau, the performance remains significantly gap to that of serial HMC. This suggests that selecting an appropriate value of $\tau$ is crucial and cannot be fully compensated for a large choice of $M$. The full data result is given in Table \ref{tab:mnist_metric_diffM}.

\begin{figure}[H]
  \centering
  \includegraphics[width=\columnwidth]{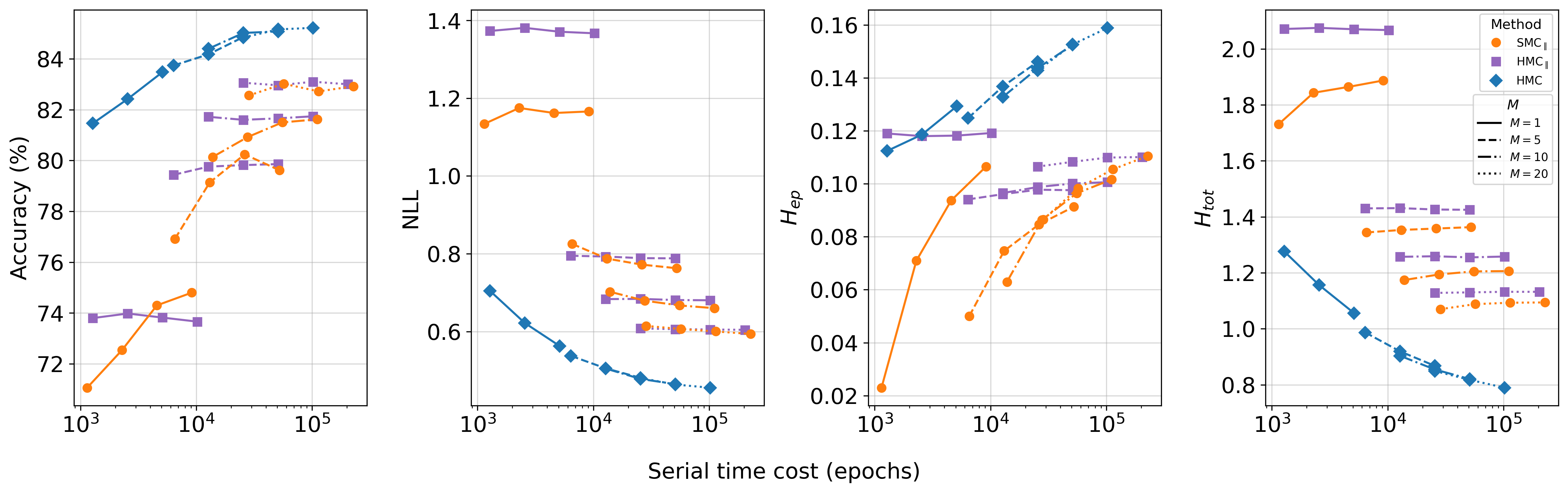}
  \caption{Comparison of SMC\(_\parallel\) (\(P\) SMC with \(N=32\), HMC\(_\parallel\) (\(NP\) chains) and HMC is with \(NP\), with \(L=1\), \(B=T=40M\) and different value of \(M\), on MNIST.}
  \label{fig:mnist_metric_diffM}
\end{figure}

\subsection{Full comparison of SMC$_\parallel$ and HMC$_\parallel$ on IMDb}
\label{app:imdb_full}

Experiments in this section are tested on the IMDb dataset with the model setting stated in Appendix \ref{app:imdb}.

\begin{figure}[H]
  \centering
  \includegraphics[width=\columnwidth]{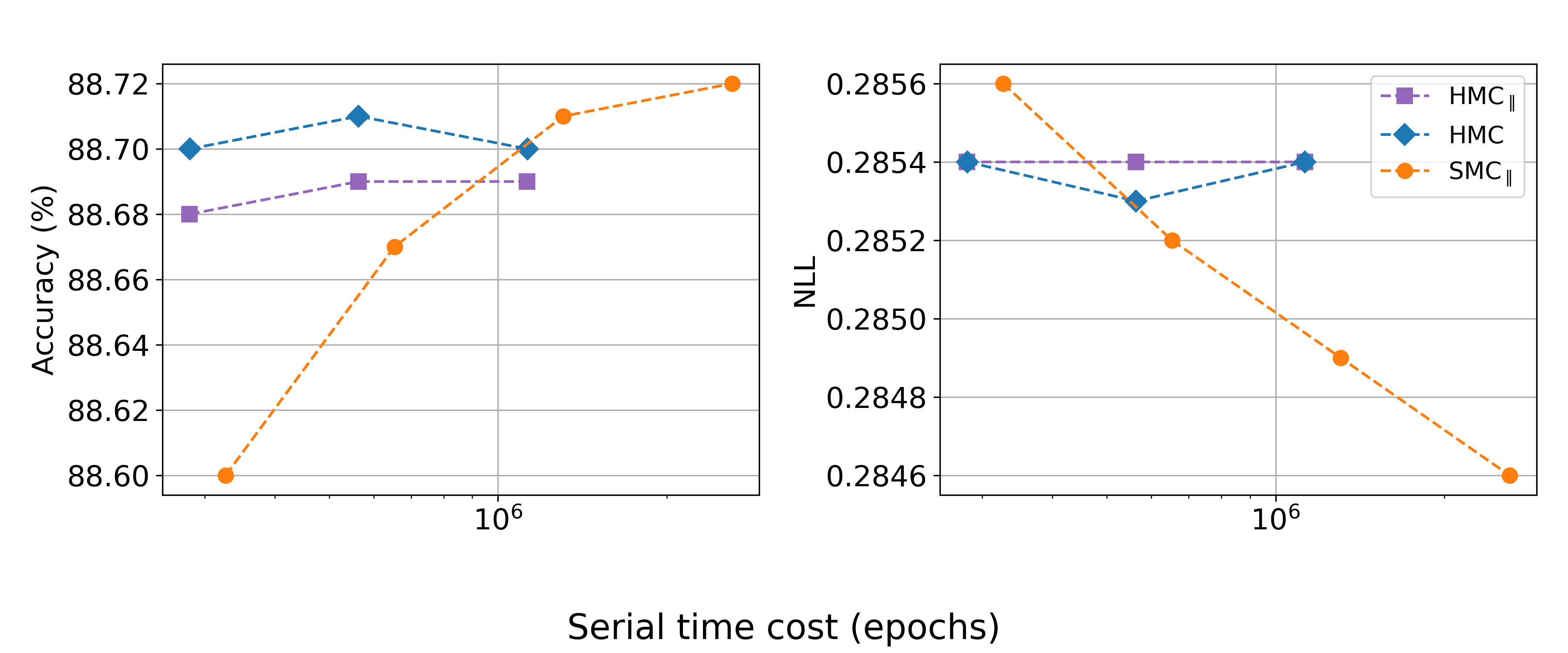}
  \caption{Comparison of SMC\(_\parallel\) (\(P\) chains with \(N=32\) particles), HMC\(_\parallel\) (\(NP\) chains) and HMC (\(NP\) particles), with fixed trajectory \(L\delta=0.1\), \(B=T=110\) and \(M=1\), on IMDb (10 realizations).}
  \label{fig:imdb_metric_fixtraj}
\end{figure}

\begin{wraptable}[16]{r}{0.6\textwidth}
\vspace{-15pt}
\centering
\scriptsize
\caption{Comparison of SMC\(_\parallel\) ($P$ SMC with 
$N=32$ particles)
and HMC\(_\parallel\) ($NP$ chains), with 
$L=1$ and $M=1,10$, 
and with fixed trajectory length \(\tau=0.1\), on IMDb.}
\label{tab:psmc_v_mcmc}
\begin{tabular}{c c | cc | cc | cc }
\toprule
\multirow{2}{*}{\(P\)} & \multirow{2}{*}{Metric} & \multicolumn{2}{c|}{\(M=1\)} & \multicolumn{2}{c|}{\(M=10\)} & \multicolumn{2}{c}{\(\tau=0.1\)} \\
\cmidrule(lr){3-4}\cmidrule(lr){5-6}\cmidrule(lr){7-8}
 & & SMC\(_\parallel\) & HMC\(_\parallel\) & SMC\(_\parallel\) & HMC\(_\parallel\) & SMC\(_\parallel\) & HMC\(_\parallel\) \\
\midrule
1 & Acc ($\uparrow$) & 79.15 & 82.47 & 86.96 & 87.11 & 88.60 & 88.68 \\
1 & NLL ($\downarrow$)& 1.56e4 & 1.6e4 & 8103 & 8095 & 7141 & 7135 \\
\midrule
2 & Acc ($\uparrow$) & 81.21 & 82.71 & 87.18 & 87.11 & 88.67 & 88.69 \\
2 & NLL ($\downarrow$)& 1.55e4 & 1.59e4 & 8046 & 8091 & 7131 & 7134 \\
\midrule
4 & Acc ($\uparrow$) & 82.52 & 82.76 & 87.24  & 87.12 & 88.71 & 88.69 \\
4 & NLL ($\downarrow$)& 1.54e4 & 1.59e4 & 8015 & 8090 & 7122 & 7134 \\
\midrule
8 & Acc ($\uparrow$) & 82.9   & 82.79  & 87.26 & 87.13 & 88.72 & -- \\
8 & NLL ($\downarrow$)& 1.53e4 & 1.59e4 & 8013 & 8087   & 7116 & -- \\
\midrule
($\downarrow$) & 
epochs &39 & 45 & 719 & 450 & 1.0e4 & 8.8e3 \\
\bottomrule
\end{tabular}
\end{wraptable}
Comparison among SMC\(_\parallel\), HMC\(_\parallel\) and HMC with different setting on $L$ and $\tau$ and increasing \(P\) is given in Figure \ref{fig:imdb_metric_fixtraj} ($\tau=0.1$) and \ref{fig:imdb_metric_diffM} ($L=1$). The full data results for these two figures are given in Table \ref{tab:imdb_metric_fixtraj} 
and \ref{tab:imdb_metric_diffM}
, respectively.
Select a few interested data from these results and generate them as Table \ref{tab:psmc_v_mcmc}, it shows (i) the methods perform comparably, and improve with $P$, (ii) HMC$_\parallel$ performs slightly better furthest from convergence $M=1$ with that gain narrowing as we approach convergence (along with the gain in $P$), (iii) the cost of the methods is comparable.
Note that averaging $P=4$ long chains of HMC with $NT$ samples and $\tau=0.1$
delivers Acc$=88.71\%$ and NLL$=7133$, bringing 
up the question of what should actually be the ``gold standard''.

\begin{figure}[H]
  \centering
  \includegraphics[width=\columnwidth]{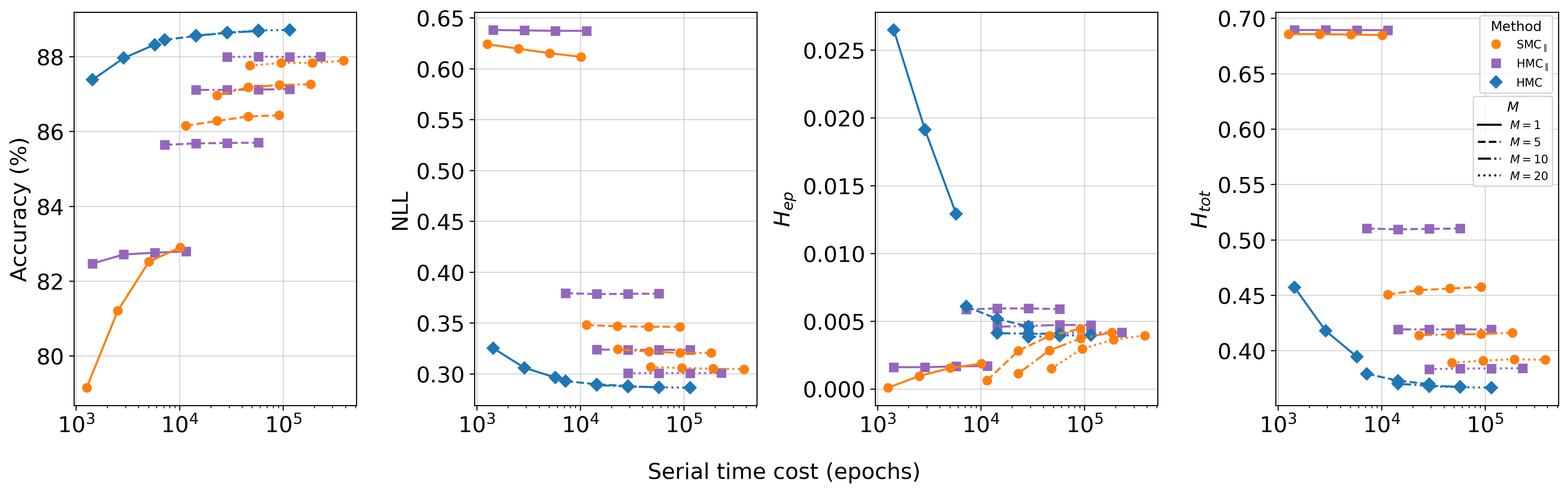}
  \caption{Comparison of SMC\(_\parallel\) (\(P\) chains with \(N=32\) particles), HMC\(_\parallel\) (\(NP\) chains) and HMC (\(P\) chains with \(N\) particles), with fixed number of leapfrog \(L=1\), \(B=T=45M\) and \(M\), on IMDb (\(5\) realizations).}
  \label{fig:imdb_metric_diffM}
\end{figure}

\subsection{Full comparison of SMC$_\parallel$ and HMC$_\parallel$ on CIFAR10}
\label{app:cifar_full}

Experiments in this section are tested on the CIFAR10 dataset with the model setting stated in Appendix \ref{app:cifar}. 

\begin{figure}[H]
  \centering
  \includegraphics[width=\columnwidth]{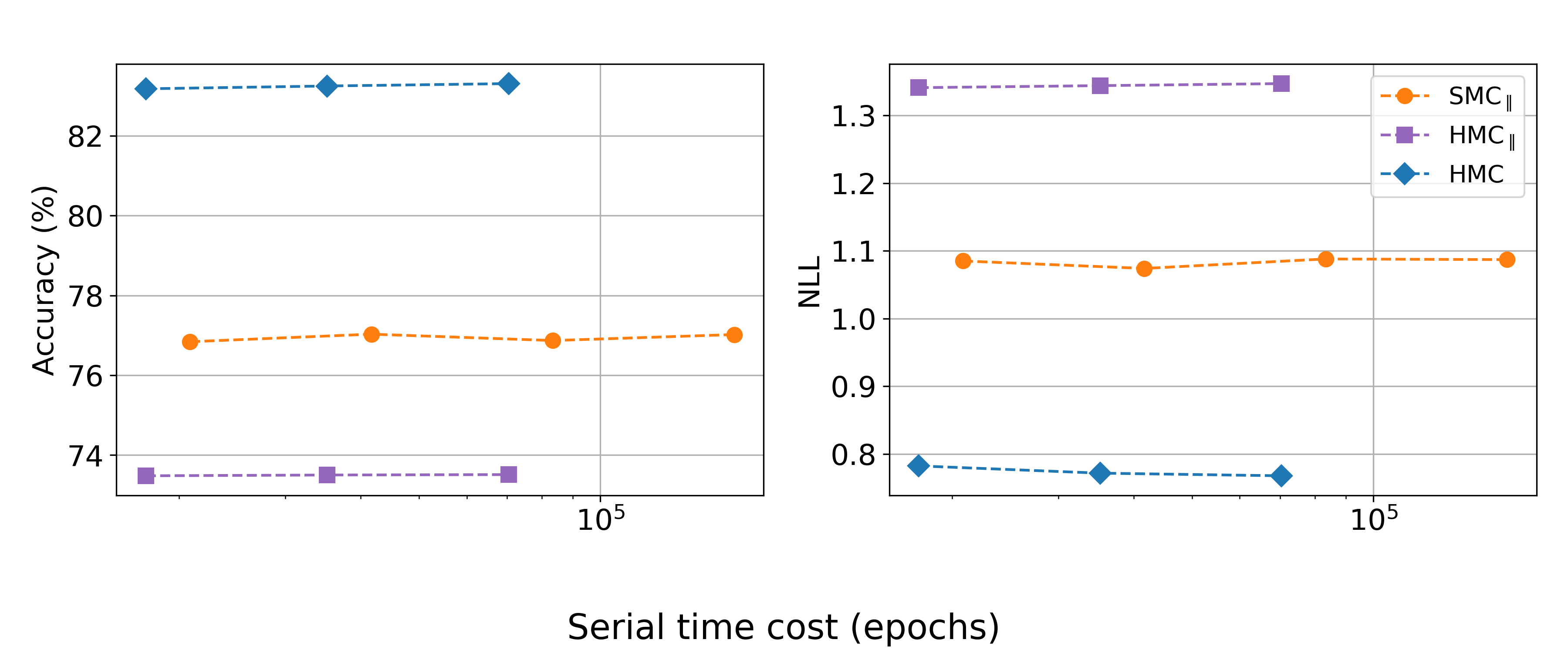}
  \caption{Comparison of SMC\(_\parallel\) (\(P\) chains with \(N=32\) particles), HMC\(_\parallel\) (\(NP\) chains) and HMC (\(P\) chains with \(N=32\) particles), with fixed number of leapfrog \(L=1\), \(B=T=550\) and \(M=5\) (\(5\) realizations).}
  \label{fig:cifar_metric_fixL}
\end{figure}

Figure \ref{fig:cifar_metric_fixL} shows the results for $L=1$, and the full data result is given in Table \ref{tab:cifar_metric_fixL}.
This figure demonstrates the failure mode when epochs have already reached $10^5$ and yet chains are still too short, highlighting a scenario where the methods need to 
be run for an unreasonably long time and are not practical.

\subsection{Further details on comparison of SMC$_\parallel$ and MCMC$_\parallel$ vs SOTA on Australian Credit data}
\label{app:other}

We compare SMC$_\parallel$ with recent
novel parallel MCMC methods from \cite{schwedes2021rao}, 
parallel MCMC, and Annealed Importance Sampling (AIS) method \cite{neal2001annealed}, with details on the AIS algorithm and optimal settings provided in subsection \ref{app:AIS}.

We evaluate the performance of methods using a Bayesian logistic regression model applied to the Australian Credit dataset, 
with $d-1=14$ covariates and  \( m = 690 \) data points. 
In this binary classification problem, 
the categorical variables are \(y_i \in \{ 0,1 \}\) for \(i=1,...,m\) 
and the design matrix is \(X = (X_1,...,X_m) \in \bbR^{m \times d}\), 
where \(X_{i,\cdot} := (1,X_{i,1},...,X_{i,d-1}) \in \bbR^d\). 
The likelihood for $\theta \in \bbR^d$ is given by \eqref{eqn:bayes_class} with $K=2$ and $D=1$, 
and the prior is \( \pi_0=\mathcal{N}(0,100 {\sf Id})\).

Since some of the compared methods lack available code or convergence results regarding MSE, we assess all methods detailed in \cite{schwedes2021rao} by comparing the empirical variance of posterior mean estimates. Figure~\ref{fig:parallelMCMC_AIS} presents the empirical variance for the proposed methods as a function of both the number of samples and the number of processors. This comparison includes PSMC, along with: Metropolis-Hastings (MH), Simplified manifold Metropolis-adjusted Langevin MH (MH SmMALA) \cite{girolami2011riemann}, Rao-Blackwellised Multiple Proposal MCMC (RB-MP-MCMC) \cite{schwedes2021rao}, Adaptive Rao-Blackwellised Multiple Proposal MCMC (ARB-MP-MCMC) \cite{schwedes2021rao}, Naive parallel MH, Naive parallel pCN and Optimal AIS \cite{neal2001annealed}. As shown in Figure~\ref{fig:parallelMCMC_AIS}, PSMC achieves performance comparable to the novel parallel MCMC methods and significantly outperforms naive parallel MCMC and Optimal AIS methods.
The data for all MCMC methods shown in Figure~\ref{fig:parallelMCMC_AIS} are sourced directly from Figure 4(a) in \cite{schwedes2021rao}.


\subsubsection{Details of AIS}
\label{app:AIS}

Annealed Importance Sampling was introduced in \cite{neal2001annealed} (AIS). Both AIS and SMC aim to estimate the posterior distribution through a sequence of tempering distributions, while SMC involves resampling steps to reselect the particle population, which distinguishes it from the single-particle approach in AIS and often results in improved efficiency and reduced variance in high-dimensional problems. The Algorithm of AIS is shown in Algorithm \ref{alg:ais}.

\begin{algorithm}[H]
   \caption{AIS sampler}
   \label{alg:ais}
\begin{algorithmic}
    \STATE \textbf{Inputs:} $\cL$, $\pi_0$, $N$.
   \STATE \(\theta^i_0 \sim \pi_0\) for \(i=1,...,N\) and \(Z^N=1\).
   \FOR{\(i=1\) {\bfseries to} \(N\) (in parallel)} 
   \FOR{\(j=1\) {\bfseries to} \(J\)}
    \STATE Define the importance weight: \(w^i_j = \cL(\theta_{j-1}^k)^{\lambda_j-\lambda_{j-1}}\).
    \STATE Store \(Z^i *= w^i_j\).
    \STATE \textbf{Mutation}: Draw \(\theta_j^i \sim \cM_j(\theta_{j-1}^i, \cdot )\).
    \ENDFOR
    \ENDFOR
    \STATE \textbf{Outputs:} $\{\theta^i=\theta_{J}^{i}\}_{i=1}^{N}$ and $Z^{N}$.
\end{algorithmic}
\end{algorithm}

AIS is fully parallelable in each sample and has consistent estimators as in the SMC. 
The estimator of AIS for the QOI, \(\varphi\), is given by
\begin{equation*}
    \hat{\varphi}_{\text{AIS}} = \frac{\sum_{i=1}^{N} \varphi(\theta^i) Z^{i} }{\sum_{i=1}^{N} Z^{i}}
\end{equation*}

The approach to choose the tempering steps is optimal and is followed from examples in Neal's paper, where 40 intermediate distributions: 4 distributions linearly spaced from \(\beta=0\) to \(\beta=0.001\), 7 distributions geometrically spaced from \(\beta=0.001\) to \(\beta=0.01\) and 29 distributions geometrically spaced from \(\beta=0.01\) to \(\beta=1\). The optimal is justified by evaluating the Effective Sample Size (ESS) of the AIS method using \(2044\) samples at each tempering step. The ESS is consistently larger than or equal to \(N/2\), maintaining an adequate particle set and ensuring a fair comparison with the PSMC method, which employs adaptive tempering to maintain ESS \(\geq N/2\). The comparison results for AIS with PSMC are shown in Figure \ref{fig:parallelMCMC_AIS}. It suggests that AIS is comparable to the standard MCMC methods, like MH, MH smMALA and the naive parallel MCMC methods, but it performs worse than PSMC or other novel parallel MCMC methods.

\section{All-inclusive data tables}

\begin{table}[h]
\footnotesize
\caption{Comparison of SMC\(_\parallel\) (\(P\) SMC with \(N\) particles) and SMC (\(NP\) samples), with fixed trajectory \(L\delta=0.1\) and \(M=1\), on MNIST (\(5\) realizations and \(\pm\) s.e. in accuracy).}
\label{tab:mnist_hpc_inter_connect}
\centering
\begin{tabular}{lll|c|c|c|c|c|c}
\toprule
$P$ & $N$ & Method & Time & Epochs & Acc. & NLL & \(H_{\sf epochs}\) & \(H_{\sf tot}\)\\
\midrule
\(1\) & $16$ &SMC\(_\parallel\) & 97.62 & 5.146e+2 & 85.00\(\pm\)0.19 & 4.744e-1 & 1.398e-1 & 7.992e-1 \\
\midrule
\(2\) & $16$ &SMC\(_\parallel\) & 97.62 & 5.146e+2 & 84.84\(\pm\)0.10 & 4.708e-1 & 1.522e-1 & 8.156e-1 \\
\midrule
\(1\) & $32$ &SMC\(_\parallel\) & 98.78 & 6.693e+2 & 84.88\(\pm\)0.19 & 4.676e-1 & 1.494e-1 & 8.028e-1 \\
\midrule
\(2\) & $32$ &SMC\(_\parallel\) & 98.78 & 6.693e+2 & 84.94\(\pm\)0.08 & 4.654e-1 & 1.538e-1 & 8.085e-1 \\
\midrule
\(4\) & $32$ &SMC\(_\parallel\) & 98.78 & 6.693e+2 & 85.14\(\pm\)0.01 & 4.626e-1 & 1.566e-1 & 8.067e-1 \\
\midrule
\(1\) & $128$ &SMC & 136.4 & 7.592e+2 & 85.40\(\pm\)0.01 & 4.578e-1 & 1.538e-1 & 7.916e-1 \\
\midrule
\(8\) & $32$ &SMC\(_\parallel\)  & 98.78 & 6.693e+2 & 85.26\(\pm\)0.01 & 4.600e-1 & 1.586e-1  & 8.066e-1 \\
\midrule
\(1\) & $256$ &SMC  & 172.5 & 8.134e+2 & 85.36\(\pm\)0.07 & 4.545e-1 & 1.592e-1 & 7.870e-1 \\
\midrule
\(16\) & $32$ &SMC\(_\parallel\) & 98.78 & 6.693e+2 & 85.30\(\pm\)0.02 & 4.609e-1 & 1.653e-1 & 8.215e-1 \\
\midrule
\(1\) & $512$ &SMC & 221.0 & 8.360e+2 & 85.42\(\pm\)0.02 & 4.548e-1 & 1.595e-1 & 7.845e-1 \\
\midrule
\(32\) & $32$ &SMC\(_\parallel\) & 98.78 & 6.693e+2 & 85.40\(\pm\)0.01 & 4.597e-1 & 1.631e-1 & 8.167e-1 \\
\midrule
\(1\) & $1024$ &SMC & 326.1 & 8.340e+2 & 85.42\(\pm\)0.04 & 4.546e-1 & 1.604e-1 & 7.877e-1  \\
\bottomrule
\end{tabular}
\end{table}

\begin{table}[h]
\scriptsize
\caption{Comparison of SMC\(_\parallel\) (\(P\) SMC with \(N=32\), HMC\(_\parallel\) (\(NP\) chains) and HMC is with \(NP\), with fixed trajectory \(L\delta=0.005\), \(B=T=900\) and \(M=20\), on MNIST (\(10\) realizations and \(\pm\) s.e. in accuracy).}
\label{tab:mnist_metric_fixtraj0005}
\centering
\begin{adjustbox}{width=\textwidth}
\begin{tabular}{l|ccccc|ccccc}
\toprule
\(P\) & \multicolumn{5}{c|}{Acc.} & \multicolumn{5}{c}{NLL}\\
\cmidrule{2-6} \cmidrule{7-11}
& SMC\(_\parallel\) & HMC\(_\parallel\) & HMC & HMC (GS) & HMC\(_\parallel\)($N=1$) & SMC\(_\parallel\) & HMC\(_\parallel\) & HMC & HMC (GS) & HMC\(_\parallel\)($N=1$)\\
\midrule
\(1\) & 82.12\(\pm\)0.19 & 83.16\(\pm\)0.13 & 84.92\(\pm\)0.13 & 85.30\(\pm\)0.09 & 78.59\(\pm\)0.36 & 6.166e-1 & 5.961e-1 & 4.757e-1 & 4.488e-1 & 7.026e-1\\
\midrule
\(2\) & 82.69\(\pm\)0.10 & 83.28\(\pm\)0.06 & 85.28\(\pm\)0.14 & - & - & 6.010e-1 & 5.934e-1 & 4.612e-1 & - & -\\
\midrule
\(4\) & 82.98\(\pm\)0.16 & 83.22\(\pm\)0.06 & 85.33\(\pm\)0.10 & - & - & 5.950e-1 & 5.923e-1 & 4.539e-1 & - & -\\
\midrule
\(8\) & 83.02\(\pm\)0.08 & - & - & - & - &5.906e-1 & - & - & - & -\\
\midrule
Epochs & 906 & 902.55 & 3.025e+4 (\(\times P\)) & 2.620e+5 & -\\
\bottomrule
\end{tabular}
\end{adjustbox}
\end{table}

\begin{table}[h]
\scriptsize
\caption{Comparison of SMC\(_\parallel\) (\(P\) SMC and \(N\) particles), HMC\(_\parallel\) (\(NP\) chains) and HMC (\(NP\) particles), with fixed trajectory \(L\delta=0.02\), \(B=T=900\) and \(M=20\), on MNIST (\(10\) realizations and \(\pm\) s.e. in accuracy).}
\label{tab:mnist_metric_fixtraj002}
\centering
\begin{adjustbox}{width=\textwidth}
\begin{tabular}{l|ccccc|ccccc}
\toprule
\(P\) & \multicolumn{5}{c|}{Acc.} & \multicolumn{5}{c}{NLL} \\
\cmidrule{2-6} \cmidrule{7-11}
& SMC\(_\parallel\) & HMC\(_\parallel\) & HMC & HMC (GS) & HMC\(_\parallel\)($N=1$)
& SMC\(_\parallel\) & HMC\(_\parallel\) & HMC & HMC (GS) & HMC\(_\parallel\)($N=1$) \\
\midrule
\(1\) & 84.93\(\pm\)0.08 & 85.30\(\pm\)0.09 & 85.17\(\pm\)0.11 & 85.52\(\pm\)0.06 & 79.99\(\pm\)0.42
& 4.800e-1 & 4.558e-1 & 4.533e-1 & 4.457e-1 & 6.081e-1 \\
\midrule
\(2\) & 84.83\(\pm\)0.04 & 85.43\(\pm\)0.10 & 85.31\(\pm\)0.10 & - & -
& 4.758e-1 & 4.540e-1 & 4.494e-1 & - & - \\
\midrule
\(4\) & 84.98\(\pm\)0.01 & 85.38\(\pm\)0.10 & - & - & -
& 4.756e-1 & 4.532e-1 & - & - & - \\
\midrule
\(8\) & 85.15\(\pm\)0.02 & - & - & - & -
& 4.749e-1 & - & - & - & - \\
\midrule
Epochs 
& 2.040e+3 & 4.637e+3 & 1.764e+5 (\(\times P\)) & 1.300e+6 & - \\
\bottomrule
\end{tabular}
\end{adjustbox}
\end{table}

\begin{table}[h]
\scriptsize
\caption{Comparison of SMC\(_\parallel\) (\(P\) SMC and \(N=32\) particles), HMC\(_\parallel\) (\(NP\) chains) and HMC (\(NP\) particles), with fixed trajectory \(L\delta=0.1\), \(B=T=900\) and \(M=20\), on MNIST (\(10\) realizations and \(\pm\) s.e. in accuracy).}
\label{tab:mnist_metric_fixtraj01}
\centering
\begin{adjustbox}{width=\textwidth}
\begin{tabular}{l|ccccc|ccccc}
\toprule
\(P\) & \multicolumn{5}{c|}{Acc.} & \multicolumn{5}{c}{NLL} \\
\cmidrule{2-6} \cmidrule{7-11}
& SMC\(_\parallel\) & HMC\(_\parallel\) & HMC & HMC (GS) & HMC\(_\parallel\)($N=1$) 
& SMC\(_\parallel\) & HMC\(_\parallel\) & HMC & HMC (GS) & HMC\(_\parallel\)($N=1$)\\
\midrule
\(1\)  & 85.53\(\pm\)0.05 & 85.20\(\pm\)0.12 & 85.35\(\pm\)0.11 & 85.58\(\pm\)0.02 & 80.04\(\pm\)0.37 
       & 4.509e-1 & 4.529e-1 & 4.549e-1 & 4.461e-1 & 6.119e-1\\
\midrule
\(2\)  & 85.42\(\pm\)0.01 & 85.46\(\pm\)0.08 & - & - & - 
       & 4.494e-1 & 4.493e-1 & -       & -       & -      \\
\midrule
\(4\)  & 85.50\(\pm\)0.09 & 85.41\(\pm\)0.09 & - & - & - 
       & 4.475e-1 & 4.483e-1 & -       & -       & -      \\
\midrule
\(8\)  & 85.50\(\pm\)0.01 & -             & - & - & - 
       & 4.471e-1 & -       & -       & -       & -      \\
\midrule
Epochs    & 1.651e+4        & 2.996e+4        & 9.418e+5 (\(\times P\)) & 6.723e+6 & - \\
\bottomrule
\end{tabular}
\end{adjustbox}
\end{table}

\begin{table}[h]
\scriptsize
\caption{SMC\(_\parallel\) (\(P\) SMC with \(N\) particles), HMC\(_\parallel\) (\(NP\) chains) and HMC (\(NP\) particles), with fixed number of leapfrog \(L=1\), \(M\) mutation steps and \(B=T=40M\), on MNIST (\(5\) realizations and \(\pm\) s.e. in accuracy).}
\label{tab:mnist_metric_diffM}
\centering
\subcaption{Acc.}
\begin{tabular}{l|ccc|ccc}
\toprule
\(P\) & \multicolumn{3}{c|}{\textbf{\(M=1\)}} & \multicolumn{3}{c}{\textbf{\(M=5\)}} \\
& SMC\(_\parallel\) & HMC\(_\parallel\) & HMC & SMC\(_\parallel\) & HMC\(_\parallel\) & HMC \\
\midrule
\(1\) & 71.06\(\pm\)3.92 & 73.80\(\pm\)0.47 & 81.46\(\pm\)0.38 & 76.92\(\pm\)0.49 & 79.44\(\pm\)0.25 & 83.74\(\pm\)0.13 \\
\midrule
\(2\) & 72.54\(\pm\)3.32 & 73.98\(\pm\)0.38 & 82.42\(\pm\)0.24 & 79.14\(\pm\)0.22 & 79.76\(\pm\)0.18 & 84.18\(\pm\)0.28 \\
\midrule
\(4\) & 74.30\(\pm\)0.54 & 73.82\(\pm\)0.19 & 83.48\(\pm\)0.16 & 80.24\(\pm\)0.27 & 79.82\(\pm\)0.16 & 84.86\(\pm\)0.11 \\
\midrule
\(8\) & 74.80\(\pm\)0.21 & 73.66\(\pm\)0.10 &  -  & 79.62\(\pm\)0.24 & 79.86\(\pm\)0.12 &  - \\
\midrule
Epochs & 35.75 & 40 & 1280 (\(\times P\)) & 204.87 & 200 & 6400 (\(\times P\)) \\ 
\bottomrule
\multicolumn{2}{c}{} \\
\toprule
 \(P\) & \multicolumn{3}{c|}{\textbf{\(M=10\)}} & \multicolumn{3}{c}{\textbf{\(M=20\)}} \\
& SMC\(_\parallel\) & HMC\(_\parallel\) & HMC & SMC\(_\parallel\) & HMC\(_\parallel\) & HMC \\
\midrule
\(1\) & 80.14\(\pm\)0.31 & 81.72\(\pm\)0.24 & 84.40\(\pm\)0.18 & 82.56\(\pm\)0.09 & 83.06\(\pm\)0.24 & 84.84\(\pm\)0.20 \\
\midrule
\(2\) & 80.92\(\pm\)0.93 & 81.60\(\pm\)0.10 & 85.02\(\pm\)0.15 & 83.02\(\pm\)0.07 & 82.96\(\pm\)0.14 & 85.16\(\pm\)0.24 \\
\midrule
\(4\) & 81.50\(\pm\)0.06 & 81.66\(\pm\)0.10 & 85.08\(\pm\)0.19 & 82.72\(\pm\)0.06 & 83.10\(\pm\)0.18 & 85.22\(\pm\)0.16 \\
\midrule
\(8\) & 81.62\(\pm\)0.18 & 81.74\(\pm\)0.11 &  -  & 82.92\(\pm\)0.03 & 83.00\(\pm\)0.10 & - \\
\midrule
Epochs & 433.50 & 400 & 12800 (\(\times P\)) & 892 & 800 & 25600 (\(\times P\)) \\
\bottomrule
\end{tabular}
\\

\subcaption{NLL}
\begin{adjustbox}{width=\textwidth}
\begin{tabular}{l|ccc|ccc|ccc|ccc}
\toprule
\(P\) & \multicolumn{3}{c|}{\textbf{\(M=1\)}} & \multicolumn{3}{c|}{\textbf{\(M=5\)}} & \multicolumn{3}{c|}{\textbf{\(M=10\)}} & \multicolumn{3}{c}{\textbf{\(M=20\)}} \\
& SMC\(_\parallel\) & HMC\(_\parallel\) & HMC & SMC\(_\parallel\) & HMC\(_\parallel\) & HMC & SMC\(_\parallel\) & HMC\(_\parallel\) & HMC & SMC\(_\parallel\) & HMC\(_\parallel\) & HMC \\
\midrule
\(1\) & 1.134e+0 & 1.373e+0 & 7.046e-1 & 8.253e-1 & 7.949e-1 & 5.375e-1 & 7.024e-1 & 6.834e-1 & 5.048e-1 & 6.145e-1 & 6.080e-1  & 4.804e-1 \\
\midrule
\(2\) & 1.175e+0 & 1.381e+0 & 6.219e-1 & 7.875e-1 & 7.929e-1 & 5.053e-1 & 6.794e-1 & 6.840e-1 & 4.781e-1 &  6.064e-1 & 6.061e-1  & 4.637e-1 \\
\midrule
\(4\) & 1.162e+0 & 1.371e+0 & 5.628e-1 & 7.721e-1 & 7.888e-1 & 4.808e-1 & 6.671e-1 & 6.810e-1 & 4.647e-1 & 6.002-1  &  6.052e-1 & 4.555e-1 \\
\midrule
\(8\) & 1.166e+0 & 1.367e+0 & - & 7.628e-1 & 7.882e-1 & - & 6.601e-1 & 6.804e-1 & - & 5.938e-1  &  6.044e-1 & - \\
\bottomrule
\end{tabular}
\end{adjustbox}

\subcaption{\(H_{\sf ep}\)}
\begin{adjustbox}{width=\textwidth}
\begin{tabular}{l|ccc|ccc|ccc|ccc}
\toprule
\(P\) & \multicolumn{3}{c|}{\textbf{\(M=1\)}} & \multicolumn{3}{c|}{\textbf{\(M=5\)}} & \multicolumn{3}{c|}{\textbf{\(M=10\)}} & \multicolumn{3}{c}{\textbf{\(M=20\)}} \\
& SMC\(_\parallel\) & HMC\(_\parallel\) & HMC & SMC\(_\parallel\) & HMC\(_\parallel\) & HMC & SMC\(_\parallel\) & HMC\(_\parallel\) & HMC & SMC\(_\parallel\) & HMC\(_\parallel\) & HMC \\
\midrule
\(1\) & 2.297e-2 & 1.190e-1 & 1.1242e-1 & 5.001e-2 & 9.410e-2 & 1.248e-1 & 6.296e-2 & 9.652e-2 & 1.328e-1 & 8.650e-2 & 1.065e-1 & 1.429e-1 \\
\midrule
\(2\) & 7.094e-2 & 1.180e-1 & 1.186e-1 & 7.465e-2 & 9.609e-2 & 1.368e-1 & 8.635e-2 & 9.878e-2 & 1.437e-1 & 9.833e-2 & 1.083e-1 & 1.526e-1 \\
\midrule
\(4\) & 9.372e-2 & 1.182e-1 & 1.293e-1 & 8.466e-2 & 9.777e-2 & 1.461e-1 & 9.641e-2 & 1.001e-1 & 1.525e-1 & 1.055e-1 & 1.099e-1 & 1.589e-1 \\
\midrule
\(8\) & 1.065e-1 & 1.192e-1 & - & 9.134e-2 & 9.758e-2 & - & 1.016e-1 & 1.006e-1 & - & 1.105e-1 & 1.101e-1 & - \\
\bottomrule
\end{tabular}
\end{adjustbox}

\subcaption{\(H_{\sf tot}\)}
\begin{adjustbox}{width=\textwidth}
\begin{tabular}{l|ccc|ccc|ccc|ccc}
\toprule
\(P\) & \multicolumn{3}{c|}{\textbf{\(M=1\)}} & \multicolumn{3}{c|}{\textbf{\(M=5\)}} & \multicolumn{3}{c|}{\textbf{\(M=10\)}} & \multicolumn{3}{c}{\textbf{\(M=20\)}} \\
& SMC\(_\parallel\) & HMC\(_\parallel\) & HMC & SMC\(_\parallel\) & HMC\(_\parallel\) & HMC & SMC\(_\parallel\) & HMC\(_\parallel\) & HMC & SMC\(_\parallel\) & HMC\(_\parallel\) & HMC \\
\midrule
\(1\) & 1.731e+0 & 2.071e+0 & 1.276e+0 & 1.344e+0 & 1.430e+0 & 9.867e-1 & 1.174e+0 & 1.257e+0 & 9.032e-1 & 1.070e+0 & 1.128e+0 & 8.490e-1 \\
\midrule
\(2\) & 1.843e+0 & 2.075e+0 & 1.156e+0 & 1.353e+0 & 1.431e+0 & 9.191e-1 & 1.194e+0 & 1.259e+0 & 8.544e-1 & 1.088e+0 & 1.130e+0 & 8.173e-1 \\
\midrule
\(4\) & 1.864e+0 & 2.070e+0 & 1.056e+0 & 1.358e+0 & 1.426e+0 & 8.673e-1 & 1.205e+0 & 1.255e+0 & 8.202e-1 & 1.093e+0 & 1.132e+0 & 7.893e-1 \\
\midrule
\(8\) & 1.887e+0 & 2.067e+0 & - & 1.363e+0 & 1.425e+0 & - & 1.206e+0 & 1.258e+0 & - & 1.094e+0 & 1.132e+0 & - \\
\bottomrule
\end{tabular}
\end{adjustbox}
\end{table}

\begin{table}[h]
\scriptsize
\caption{Comparison of SMC\(_\parallel\) (\(P\) chains with \(N=32\) particles), HMC\(_\parallel\) (\(NP\) chains) and HMC (\(NP\) particles), with fixed trajectory \(L\delta=0.1\), \(B=T=110\) and \(M=1\), on IMDb (10 realizations and \(\pm\) s.e.\ in accuracy).}
\label{tab:imdb_metric_fixtraj}
\centering
  \begin{tabular}{l|cccc|cccc}
    \toprule
    \(P\) & \multicolumn{4}{c|}{Acc.} & \multicolumn{4}{c}{NLL}\\
     & SMC\(_\parallel\) & HMC\(_\parallel\) & HMC & HMC (GS)
     & SMC\(_\parallel\) & HMC\(_\parallel\) & HMC & HMC (GS)\\
    \midrule
    1 & 88.60\(\pm\)0.0014 & 88.68\(\pm\)0.01 & 88.70\(\pm\)0.01 & 88.68\(\pm\)0.01
      & 2.856e-1 & 2.854e-1 & 2.854e-1 & 2.853e-1\\
    \midrule
    2 & 88.67\(\pm\)0.0007 & 88.69\(\pm\)0.01 & 88.71\(\pm\)0.01 & –
      & 2.852e-1 & 2.854e-1 & 2.853e-1 & –\\
    \midrule
    4 & 88.71\(\pm\)0.0012 & 88.69\(\pm\)0.01 & 88.70\(\pm\)0.01 & –
      & 2.849e-1 & 2.854e-1 & 2.854e-1 & –\\
    \midrule
    8 & 88.72\(\pm\)0.0006 & – & – & –
      & 2.846e-1 & – & – & –\\
    \midrule
    Epochs\ 
      & 1.021e+4 & 8.806e+3 & 2.820e+5 (\(\times P\)) & 5.164e+4
      & \multicolumn{4}{c}{} \\
    \bottomrule
\multicolumn{2}{c}{} \\
    \toprule
    \(P\) & \multicolumn{4}{c|}{\(H_{\sf ep}\)} & \multicolumn{4}{c}{\(H_{\sf tot}\)}\\
     & SMC\(_\parallel\) & HMC\(_\parallel\) & HMC & HMC (GS)
     & SMC\(_\parallel\) & HMC\(_\parallel\) & HMC & HMC (GS)\\
    \midrule
    1 & 2.904e-3 & 3.788e-3 & 3.760e-3 & 3.866e-3
      & 3.601e-1 & 3.615e-1 & 3.616e-1 & 3.616e-1\\
    \midrule
    2 & 3.658e-3 & 3.835e-3 & 3.787e-3 & –
      & 3.619e-1 & 3.617e-1 & 3.614e-1 & –\\
    \midrule
    4 & 4.181e-3 & 3.868e-3 & 3.840e-3 & –
      & 3.623e-1 & 3.617e-1 & 3.617e-1 & –\\
    \midrule
    8 & 4.368e-3 & – & – & –
      & 3.627e-1 & – & – & –\\
    \bottomrule
  \end{tabular}
\end{table}

\begin{table}[h]
\scriptsize
\caption{Comparison of SMC\(_\parallel\) (\(P\) chains with \(N=32\) particles), HMC\(_\parallel\) (\(NP\) chains) and HMC (\(P\) chains with \(N=32\) particles), with fixed number of leapfrog \(L=1\), \(B=T=550\) and \(M=5\), on CIFAR10 (\(5\) realizations and \(\pm\) s.e. in accuracy).}
\label{tab:cifar_metric_fixL}
\centering
  \begin{tabular}{l|cccc|cccc}
    \toprule
    \(P\) & \multicolumn{4}{c|}{Acc.} & \multicolumn{4}{c}{NLL} \\
          & SMC\(_\parallel\) & HMC\(_\parallel\) & HMC          & HMC (GS)       & SMC\(_\parallel\) & HMC\(_\parallel\) & HMC          & HMC (GS)       \\
    \midrule
    \(1\) & 76.84\(\pm\)0.28 & 73.48\(\pm\)0.11 & 83.18\(\pm\)0.01 & 81.70\(\pm\)0.21 & 1.085e+0 & 1.341e+0 & 7.823e-1 & 8.318e-1 \\
    \midrule
    \(2\) & 77.03\(\pm\)0.16 & 73.5\(\pm\)0.03  & 83.25\(\pm\)0.01 & –                & 1.074e+0 & 1.344e+0 & 7.716e-1 & –        \\
    \midrule
    \(4\) & 76.87\(\pm\)0.15 & 73.51\(\pm\)0.03 & 83.31\(\pm\)0.02 & –                & 1.088e+0 & 1.347e+0 & 7.677e-1 & –        \\
    \midrule
    \(8\) & 77.02\(\pm\)0.08 & –                & –                & –                & 1.087e+0 & –        & –         & –        \\
    \midrule
    Epochs    & 6.515e+2         & 550              & 1.76e+4 (\(\times P\)) & 1999           & & & & \\
    \bottomrule

\multicolumn{2}{c}{} \\

    \toprule
    \(P\) & \multicolumn{4}{c|}{\(H_{\sf ep}\)} & \multicolumn{4}{c}{\(H_{\sf tot}\)} \\
          & SMC\(_\parallel\) & HMC\(_\parallel\) & HMC        & HMC (GS)     & SMC\(_\parallel\) & HMC\(_\parallel\) & HMC        & HMC (GS)     \\
    \midrule
    \(1\) & 3.535e-4  & 1.002e-2 & 1.429e-1 & 2.631e-3 & 1.669e+0 & 1.915e+0 & 1.409e+0 & 1.428e+0 \\
    \midrule
    \(2\) & 1.779e-3  & 9.885e-3 & 6.952e-3 & –        & 1.660e+0 & 1.916e+0 & 1.387e+0 & –        \\
    \midrule
    \(4\) & 2.884e-3  & 9.207e-3 & 5.294e-3 & –        & 1.676e+0 & 1.918e+0 & 1.380e+0 & –        \\
    \midrule
    \(8\) & 3.386e-3  & –        & –        & –        & 1.677e+0 & –        & –        & –        \\
    \bottomrule
    \end{tabular}
\end{table}

\begin{table}[h]
\scriptsize
\caption{Comparison of SMC\(_\parallel\) (\(P\) chains with \(N=32\) particles), HMC\(_\parallel\) (\(NP\) chains) and HMC (\(P\) chains with \(N\) particles), with fixed number of leapfrog \(L=1\), \(B=T=45M\) and \(M\), on IMDb (\(5\) realizations and \(\pm\) s.e. in accuracy).} 
\label{tab:imdb_metric_diffM}
\centering
\subcaption{Acc.}
\begin{tabular}{l|ccc|ccc}
\toprule
\(P\) & \multicolumn{3}{c|}{\(M=1\)} & \multicolumn{3}{c}{\(M=5\)} \\
& SMC\(_\parallel\) & HMC\(_\parallel\) & HMC & SMC\(_\parallel\) & HMC\(_\parallel\) & HMC \\
\midrule
\(1\) & 79.15\(\pm\)0.32 & 82.47\(\pm\)0.15 & 87.38\(\pm\)0.14 & 86.15\(\pm\)0.013 & 85.64\(\pm\)0.01 & 88.45\(\pm\)0.03 \\
\midrule
\(2\) & 81.21\(\pm\)0.07 & 82.71\(\pm\)0.05 & 87.96\(\pm\)0.09 & 86.28\(\pm\)0.013 & 85.68\(\pm\)0.02 & 88.55\(\pm\)0.02 \\
\midrule
\(4\) & 82.52\(\pm\)0.14 & 82.76\(\pm\)0.04 & 88.32\(\pm\)0.05 & 86.40\(\pm\)0.005 & 85.69\(\pm\)0.03 & 88.64\(\pm\)0.02 \\
\midrule
\(8\) & 82.90\(\pm\)0.07 & 82.79\(\pm\)0.05 &  -  & 86.43\(\pm\)0.007 & 85.70\(\pm\)0.007 &  -  \\
\midrule
Epochs & 3.955e+1 & 45 & 1440 (\(\times P\)) & 3.581e+2 & 225 & 7200 (\(\times P\)) \\
\bottomrule

\multicolumn{2}{c}{} \\

\toprule
\(P\) & \multicolumn{3}{c|}{\(M=10\)} & \multicolumn{3}{c}{\(M=20\)} \\
& SMC\(_\parallel\) & HMC\(_\parallel\) & HMC & SMC\(_\parallel\) & HMC\(_\parallel\) & HMC \\
\midrule
\(1\) & 86.96\(\pm\)0.005 & 87.11\(\pm\)0.02 & 88.56\(\pm\)0.02 & 87.76\(\pm\)0.001 & 87.99\(\pm\)0.01 & 88.64\(\pm\)0.009 \\
\midrule
\(2\) & 87.18\(\pm\)0.003 & 87.11\(\pm\)0.01 & 88.64\(\pm\)0.01 & 87.83\(\pm\)0.005 & 88.00\(\pm\)0.004 & 88.70\(\pm\)0.02 \\
\midrule
\(4\) & 87.24\(\pm\)0.0006 & 87.12\(\pm\)0.009 & 88.68\(\pm\)0.006 & 87.83\(\pm\)0.003 & 87.99\(\pm\)0.007 & 88.71\(\pm\)0.02 \\
\midrule
\(8\) & 87.26\(\pm\)0.0007 & 87.13\(\pm\)0.008 &  -  &  87.89\(\pm\)0.0003  & 88.00\(\pm\)0.009 &  - \\
\midrule
Epochs & 7.185e+2 & 450 & 14400 (\(\times P\)) & 1.495e+3 & 900 & 28800 (\(\times P\)) \\ 
\bottomrule
\end{tabular}

\subcaption{NLL}
\begin{adjustbox}{width=\textwidth}
\begin{tabular}{l|ccc|ccc|ccc|ccc}
\toprule
\(P\) & \multicolumn{3}{c|}{\(M=1\)} & \multicolumn{3}{c|}{\(M=5\)} & \multicolumn{3}{c|}{\(M=10\)} & \multicolumn{3}{c}{\(M=20\)} \\
& SMC\(_\parallel\) & HMC\(_\parallel\) & HMC & SMC\(_\parallel\) & HMC\(_\parallel\) & HMC & SMC\(_\parallel\) & HMC\(_\parallel\) & HMC & SMC\(_\parallel\) & HMC\(_\parallel\) & HMC \\
\midrule
\(1\) & 6.240e-1 & 6.380e-1 & 3.252e-1 & 3.480e-1 & 3.792e-1 & 2.929e-1 & 3.241e-1 & 3.238e-1 & 2.888e-1 & 3.066e-1 & 3.006e-1 & 2.874e-1 \\
\midrule
\(2\) & 6.196e-1 & 6.376e-1 & 3.059e-1 & 3.467e-1 & 3.784e-1 & 2.896e-1 & 3.218e-1 & 3.236e-1 & 2.876e-1 & 3.056e-1 & 3.007e-1 & 2.867e-1 \\
\midrule
\(4\) & 6.152e-1 & 6.372e-1 & 2.963e-1 & 3.462e-1 & 3.786e-1 & 2.880e-1 & 3.206e-1 & 3.236e-1 & 2.867e-1 & 3.050e-1 & 3.007e-1 & 2.863e-1 \\
\midrule
\(8\) & 6.116e-1 & 6.372e-1 & –       & 3.463e-1 & 3.788e-1 & –       & 3.205e-1 & 3.235e-1 & –       & 3.046e-1 & 3.008e-1 & –      \\
\bottomrule
\end{tabular}
\end{adjustbox}

\subcaption{\(H_{\sf ep}\)}
\begin{adjustbox}{width=\textwidth}
\begin{tabular}{l|ccc|ccc|ccc|ccc}
\toprule
\(P\) & \multicolumn{3}{c|}{\(M=1\)} & \multicolumn{3}{c|}{\(M=5\)} & \multicolumn{3}{c|}{\(M=10\)} & \multicolumn{3}{c}{\(M=20\)} \\
& SMC\(_\parallel\) & HMC\(_\parallel\) & HMC & SMC\(_\parallel\) & HMC\(_\parallel\) & HMC & SMC\(_\parallel\) & HMC\(_\parallel\) & HMC & SMC\(_\parallel\) & HMC\(_\parallel\) & HMC\\
\midrule
\(1\) & 8.420e-5 & 1.598e-3 & 2.647e-2 & 6.250e-4 & 5.858e-3 & 6.078e-3 & 1.142e-3 & 4.581e-3 & 4.106e-3 & 1.496e-3 & 4.039e-3 & 3.821e-3 \\
\midrule
\(2\) & 9.448e-4 & 1.601e-3 & 1.911e-2 & 2.803e-3 & 5.935e-3 & 5.153e-3 & 2.840e-3 & 4.645e-3 & 4.071e-3 & 2.960e-3 & 4.116e-3 & 3.929e-3 \\
\midrule
\(4\) & 1.538e-3 & 1.657e-3 & 1.291e-2 & 3.914e-3 & 5.932e-3 & 4.612e-3 & 3.728e-3 & 4.715e-3 & 4.055e-3 & 3.637e-1 & 4.152e-3 & 3.971e-3 \\
\midrule
\(8\) & 1.860e-3 & 1.679e-3 & - & 4.434e-3 & 5.884e-3 & - & 4.159e-3 & 4.707e-3 & - & 3.934e-3 & 4.159e-3 & - \\
\bottomrule
\end{tabular}
\end{adjustbox}

\subcaption{\(H_{\sf tot}\)}
\begin{adjustbox}{width=\textwidth}
\begin{tabular}{l|ccc|ccc|ccc|ccc}
\toprule
\(P\) & \multicolumn{3}{c|}{\(M=1\)} & \multicolumn{3}{c|}{\(M=5\)} & \multicolumn{3}{c|}{\(M=10\)} & \multicolumn{3}{c}{\(M=20\)} \\
& SMC\(_\parallel\) & HMC\(_\parallel\) & HMC & SMC\(_\parallel\) & HMC\(_\parallel\) & HMC & SMC\(_\parallel\) & HMC\(_\parallel\) & HMC & SMC\(_\parallel\) & HMC\(_\parallel\) & HMC\\
\midrule
\(1\) & 6.857e-1 & 6.894e-1 & 4.571e-1 & 4.507e-1 & 5.102e-1 & 3.791e-1 & 4.139e-1 & 4.192e-1 & 3.700e-1 & 3.890e-1 & 3.833e-1 & 3.678e-1 \\
\midrule
\(2\) & 6.857e-1 & 6.894e-1 & 4.179e-1 & 4.545e-1 & 5.095e-1 & 3.728e-1 & 4.146e-1 & 4.191e-1 & 3.683e-1 & 3.910e-1 & 3.838e-1 & 3.670e-1 \\
\midrule
\(4\) & 6.853e-1 & 6.893e-1 & 3.945e-1 & 4.561e-1 & 5.099e-1 & 3.698e-1 & 4.149e-1 & 4.193e-1 & 3.673e-1 & 3.922e-1 & 3.839e-1 & 3.665e-1 \\
\midrule
\(8\) & 6.848e-1 & 6.893e-1 & - & 4.573e-1 & 5.102e-1 & - & 4.162e-1 & 4.190e-1 & - & 3.917e-1 & 3.841e-1 & - \\
\bottomrule
\end{tabular}
\end{adjustbox}
\end{table}

\end{document}